\documentclass{article}
\usepackage[final,main]{neurips_2026}

\usepackage[utf8]{inputenc}
\usepackage[T1]{fontenc}
\usepackage{hyperref}
\usepackage{url}
\usepackage{booktabs}
\usepackage{amsfonts}
\usepackage{amsmath}
\usepackage{nicefrac}
\usepackage{microtype}
\usepackage{xcolor}
\usepackage{graphicx}
\usepackage{subcaption}
\usepackage{multirow}
\usepackage{float}
\usepackage{enumitem}
\usepackage{pifont}
\usepackage{makecell}
\usepackage[toc,page,header]{appendix}
\usepackage{minitoc}

\makeatletter
\def\addcontentsline#1#2#3{%
  \addtocontents{#1}{%
    \protect\contentsline{#2}{#3}{\thepage}{\@currentHref}%
  }%
}
\makeatother

\newcommand{\method}{\textsc{Probe\&Prefill}}
\newcommand{\benchmark}{\textsc{When2Tool}}

\title{LLM Agents Already Know When to Call Tools - \\ Even Without Reasoning}

\author{%
  Chung-En Sun \quad
  Linbo Liu \quad
  Ge Yan\thanks{Equal contribution.} \quad
  Zimo Wang\footnotemark[1] \quad
  Tsui-Wei Weng \\
  University of California, San Diego \\
  \texttt{\{cesun, linbo, geyan, zimowang, lweng\}@ucsd.edu}
}

\begin{document}

\maketitle

\doparttoc
\faketableofcontents

\begin{abstract}
Tool-augmented LLM agents tend to call tools indiscriminately, even when the model can answer directly. Each unnecessary call wastes API fees and latency, yet no existing benchmark systematically studies \emph{when} a tool call is actually needed. We propose \benchmark{}, a benchmark of 18 environments (15 single-hop, 3 multi-hop) spanning three categories of tool necessity --- computational scale, knowledge boundaries, and execution reliability --- each with controlled difficulty levels that create a clear decision boundary between tool-necessary and tool-unnecessary tasks. We evaluate two families of training-free baselines: Prompt-only (varying the prompt to discourage unnecessary calls) and Reason-then-Act (requiring the model to reason about tool necessity before acting). Both provide limited control: Prompt-only suppresses necessary calls alongside unnecessary ones, and Reason-then-Act still incurs a disproportionate accuracy cost on hard tasks. To understand \emph{why} these baselines fail, we probe the models' hidden states and find that tool necessity is linearly decodable from the pre-generation representation with AUROC 0.89--0.96 across six models, substantially exceeding the model's own verbalized reasoning. This reveals that \textit{models already know when tools are needed, but fail to act on this knowledge during generation}. Building on this finding, we propose \method{}, which uses a lightweight linear probe to read the hidden-state signal and prefills the model's response with a steering sentence. Across all models tested, \method{} reduces tool calls by 38\% with only 1.7\% accuracy loss, while the best baseline at comparable accuracy only reduces 4.2\% of tool calls, or achieves a similar tool call reduction but incurs a 5$\times$ higher accuracy loss. On the real-world Search-o1 agentic benchmark, \method{} reduces API calls by up to 56\% without accuracy degradation. Our code is available at: \textsf{{\small \href{https://github.com/Trustworthy-ML-Lab/when2tool}{https://github.com/Trustworthy-ML-Lab/when2tool}}}
\end{abstract}

\section{Introduction}
\label{sec:intro}

Large language models have demonstrated remarkable capabilities across a wide range of tasks, such as deep research~\citep{shao2025dr, openai2025deep}, software engineering~\citep{jimenez2023swe, liu2025migrationbench, merrill2026terminal}, mathematical reasoning~\citep{sun2025refine}, search and retrieval~\citep{jin2025search}, and user interaction~\citep{yao2024tau, barres2025tau}. Recently, the agentic paradigm has further extended these capabilities by equipping LLMs with external tools for complex planning, multi-step problem solving, and real-world interactions~\citep{schick2023toolformer, qin2023toolllm, patil2024gorilla}. However, current tool-augmented agents tend to call tools indiscriminately, even when the model already possesses the ability to answer directly. A central design question in these systems is therefore: \emph{when should the model call a tool versus solve the task directly?} In many cases, tool calls are unnecessary: an agent does not need to launch a web search or RAG pipeline to answer ``what year did humans land on the Moon?. Each unnecessary call wastes API fees, costs that compound rapidly when an agent makes dozens of decisions per session at deployment scale.

Recent work has begun to address tool-call efficiency, but existing approaches either target a different problem or bypass the question of \emph{why} models overcall. \citet{wu2025joint} reduces redundant calls by jointly refining agent instructions and tool descriptions, but focus on improving calls that are already needed, not on deciding whether a call is necessary. \citet{xu2025alignment} study when to skip tools entirely, but evaluate with oracle tools that simply return the correct answer upon invocation, and rely on SFT to modify behavior without understanding why the model overcalls. This leaves a fundamental question unanswered: \textit{do models overcall because they lack the information to decide, or because they fail to act on information they already have}?

Motivated by these limitations, we propose \benchmark{} (Section~\ref{sec:benchmark}), a benchmark designed to study the tool-call decision in a setting that closely mirrors real-world agent deployments. \benchmark{} comprises 18 environments (15 single-hop and 3 multi-hop), each providing tools that the model must invoke with correctly formatted arguments and whose responses require parsing, matching the interaction pattern of real APIs. We identify three categories of situations where an agent must decide whether to use a tool, covering what we believe are the major real-world scenarios: (1)~\emph{``Can I compute this?''}: the model understands the operation, but the operands may exceed what it can compute reliably (e.g. $12+7$ is trivial; trillion-scale multiplication is not); (2)~\emph{``Do I know this?''}: the answer may or may not exist in the model's parameters (e.g. the capital of France is common knowledge; an obscure historical date may not be); and (3)~\emph{``Can I execute this reliably?''}: the model knows the rules, but mentally tracing many sequential steps is error-prone (e.g. predicting \texttt{print(2+3)} is easy; tracing deep recursion might not). Each environment has three difficulty levels: \emph{easy} (most models can reliably solve without tools), \emph{medium} (the decision boundary where models sometimes succeed and sometimes fail), and \emph{hard} (most models cannot succeed without tools). Using \benchmark{}, we systematically evaluate two training-free baselines: Prompt-only, which varies the system prompt to discourage unnecessary calls, and Reason-then-Act, which asks the model to explicitly reason about tool necessity before acting. We find that both provide limited control over tool-call decisions, with hard tasks paying a high accuracy cost for each saved call (Section~\ref{sec:behavioral}).

Given the limited ability of prompting and reasoning, we ask a deeper question: \textit{does the model internally encode information about whether a tool is needed?} To investigate, we probe the model's hidden representations (Section~\ref{sec:probing}). We extract the hidden state at the last input token and train a simple linear classifier to predict whether a tool call is necessary. Surprisingly, the probe achieves AUROC above 0.9 across models. This reveals that \textbf{the model already encodes a clean signal about tool necessity in its hidden state}, but the generation process fails to translate it into calibrated decisions. Notably, even for models that completely fail under Reason-then-Act, the probe still extracts a strong signal, demonstrating that representation-level knowledge exists independently of the model's ability to express it through text.

Building on this finding, we propose \textbf{\method{}} (Section~\ref{sec:method}). We train a lightweight linear probe on the hidden states with binary tool-necessity labels. At inference time, the probe reads the hidden state and prefills the model's response with a short steering sentence (e.g., ``I can solve this directly without using a tool'' or ``I need to use a tool for this question''). The model then continues generating from this prefill. By adjusting the probe's decision threshold, \method{} provides smooth, fine-grained control over the accuracy--efficiency tradeoff. Across all tested models, \method{} outperforms every Prompt-only and Reason-then-Act baseline, and transfers well between tasks. It reduces unnecessary tool calls while preserving accuracy on hard tasks, requiring only a simple linear prediction and no additional reasoning from the model.

Our contributions are three-fold, progressing systematically from benchmark design and failure analysis, to a proposed mitigation method:
\begin{itemize}[leftmargin=*]
    \item \textbf{Benchmark:} We design \textbf{\benchmark{}}, the first benchmark for studying tool-call decisions. It comprises 18 environments (15 single-hop, 3 multi-hop) across three categories of tool necessity with controlled difficulty levels, totaling 1{,}080 training and 2{,}700 test tasks.
    \item \textbf{Failure Analysis:} We evaluate Prompt-only and Reason-then-Act baselines, revealing that both provide limited and coarse control over tool-call decisions, with hard tasks paying a disproportionate accuracy cost for each saved call.
    \item \textbf{Discovery \& Mitigation:} We probe pre-generation hidden states and find that tool necessity is linearly decodable. We therefore exploit this signal and propose \textbf{\method{}}, a lightweight method ($<$1ms overhead) that prefills the model's response based on the probe prediction. \textbf{\method{}} reduces tool calls by 38\% with only 1.7\% accuracy loss, while the best baselines either reduce tool calls by only 4.2\% at comparable accuracy (i.e. 9$\times$ less efficient) or suffer 5$\times$ more accuracy loss for a similar reduction. Furthermore, our method generalizes to real-world agentic search, reducing API calls by up to 56\% on the Search-o1 agentic benchmark \citep{li2025search}.
\end{itemize}

\begin{figure*}[t!]
\centering
\includegraphics[width=\textwidth]{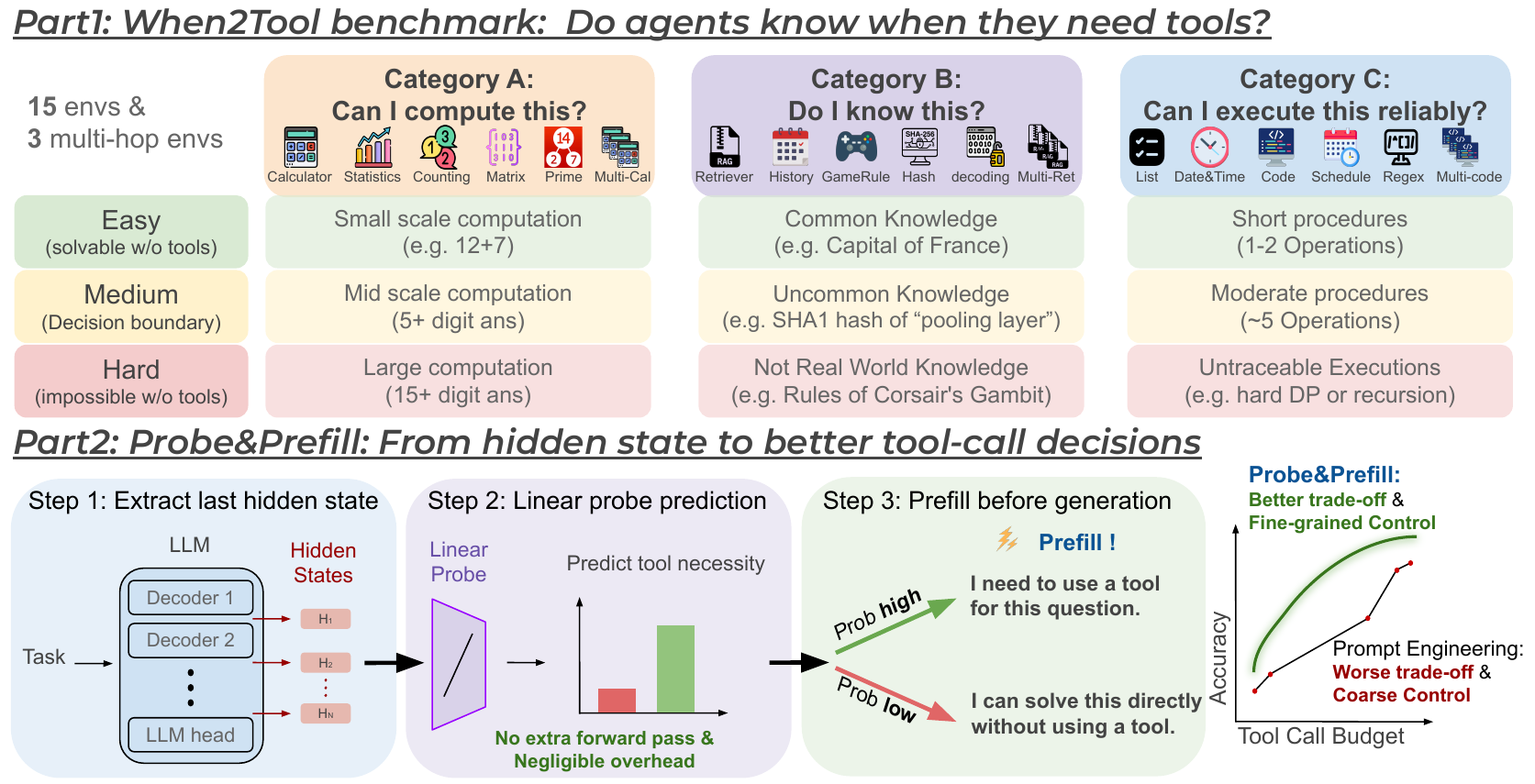}
\vspace{-15pt}
\caption{Overview. \textbf{Part 1}: We design \benchmark{} for studying whether LLM agents know when they need tools, spanning 15 single and 3 multi-hop environments across three categories, each with three difficulty levels. \textbf{Part 2}: \method{} reads the model's hidden state via a linear probe and prefills a steering sentence to guide the tool-call decision, achieving better tradeoffs.}
\vspace{-10pt}
\label{fig:overview}
\end{figure*}
\section{\benchmark{}: A Benchmark for Tool-Call Decisions}
\label{sec:benchmark}

To systematically study how LLM agents decide whether to call a tool, we build \benchmark{}, a controlled benchmark of 18 tool-use environments (15 single-hop and 3 multi-hop) spanning three categories of self-assessment. Existing benchmarks evaluate \emph{whether} models can use tools correctly~\citep{zhuang2023toolqa, li2023api}, assuming every task requires a tool. \benchmark{} instead tests \emph{whether the model knows when a tool is needed}: tasks range from those the model generally can solve directly to those that are impossible without a tool. The benchmark includes 15 single-hop environments and 3 multi-hop environments (requiring a chain of 3 dependent tool calls where each step's output is the next step's input). Table~\ref{tab:benchmark_compare} summarizes the key differences.

\subsection{Three categories of tool necessity}

In real-world agent deployments, the decision of whether to call a tool generally falls into three categories, each requiring the model to assess a different aspect of its own capability. We design 5 single-hop environments and 1 multi-hop environment for each category.
\vspace{-5pt}
\paragraph{Category A: ``Can I compute this?'' (Computational scale.)}
These environments test whether the model can assess the limits of its own mental arithmetic. The model understands the operation in every case; the question is whether the numbers involved exceed what it can compute reliably. At easy difficulty, the scale is small enough that the model can compute directly (e.g., $235 \times 48$); at hard difficulty, the operands grow to a scale that guarantees failure without a tool (e.g., trillion-scale arithmetic, $5\times 5$ determinants, $C(80,40)$). A well-calibrated agent should recognize the boundary and call the tool only when scale demands it.
\vspace{-5pt}
\paragraph{Category B: ``Do I know this?'' (Knowledge boundary.)}
These environments test whether the model can assess what information exists in its own parameters. The model must judge whether it possesses the factual knowledge needed to answer, a fundamentally different self-assessment from computational feasibility. At easy difficulty, tasks query widely known facts (e.g., the capital of France); at hard difficulty, we use fictional entities, invented events, and custom algorithms that cannot exist in any training data, guaranteeing that the model must consult the tool to get the answer. A well-calibrated agent should recognize the boundary between what it knows and what it does not.
\vspace{-5pt}
\paragraph{Category C: ``Can I execute this reliably?'' (Execution tracking.)}
These environments test whether the model can assess its own reliability when tracing sequential procedures. Unlike Category A or B, the model knows the rules of execution, and has all the information needed to produce the answer. The question is whether it can execute the steps faithfully without accumulating errors. At easy difficulty, the procedure is short enough to trace mentally (e.g., predicting the output of \texttt{print(2+3)}, checking two meetings for overlap); at hard difficulty, the procedure involves enough steps that mental execution becomes error-prone (e.g., tracing a 20-iteration dynamic programming algorithm, finding free slots across 10+ meetings). A well-calibrated agent should recognize when the execution trace exceeds its reliable tracking capacity.

\subsection{Difficulty as the decision boundary}

Each environment has three difficulty levels that control where the tool-call decision boundary falls:
\vspace{-5pt}
\begin{itemize}[leftmargin=*]
    \item \textbf{Easy}: The model can mostly solve without tools. These tasks test whether the model \emph{overcalls}, invoking tools when it does not need them.
    \item \textbf{Medium}: The decision boundary where most models sometimes succeed and sometimes fail without tools. This is where calibrated decision-making matters most.
    \item \textbf{Hard}: The model almost cannot succeed without tools. These tasks test whether the model can recognize the limits of its own capability and call the tool when it genuinely needs one.
\end{itemize}
\vspace{-5pt}
We validate these difficulty assignments empirically by running all tasks in a no-tool setting where the model is forced to answer directly (Table~\ref{tab:difficulty_validation}, Appendix~\ref{app:envs}).

In total, \benchmark{} contains 1{,}080 training tasks and 2{,}700 test tasks: 900 train / 2{,}250 test for the 15 single-hop environments, plus 180 train / 450 test for the 3 multi-hop environments (each with 3 difficulties $\times$ 20 or 50 tasks per split). Full environment details, tool descriptions, and example tasks at each difficulty level are provided in Appendix~\ref{app:envs}.

\begin{table}[h!]
\vspace{-5pt}
\caption{Comparison with existing tool-use benchmarks. \benchmark{} is the first to evaluate the tool-call decision with controlled difficulty, multi-hop tasks, and zero API cost.}
\label{tab:benchmark_compare}
\centering
\small
\setlength{\tabcolsep}{4pt}
\begin{tabular}{lccccc}
\toprule
Benchmark & \makecell{Tool-call\\decision} & \makecell{Difficulty\\levels} & \makecell{Multi-hop\\tasks} & \makecell{Realistic\\tool I/O} & \makecell{Zero\\API cost} \\
\midrule
Toolformer~\citep{schick2023toolformer}  & \ding{55} & \ding{55} & \ding{55} & \ding{55} & \ding{51} \\
ToolLLM~\citep{qin2023toolllm}          & \ding{55} & \ding{55} & \ding{51} & \ding{51} & \ding{55} \\
Gorilla~\citep{patil2024gorilla}         & \ding{55} & \ding{55} & \ding{55} & \ding{51} & \ding{55} \\
API-Bank~\citep{li2023api}               & \ding{55} & \ding{55} & \ding{51} & \ding{51} & \ding{55} \\
ToolQA~\citep{zhuang2023toolqa}          & \ding{55} & \ding{55} & \ding{51} & \ding{51} & \ding{55} \\
BFCL~\citep{patil2025berkeley}           & \ding{55} & \ding{55} & \ding{51} & \ding{51} & \ding{55} \\
\citet{xu2025alignment}                  & \ding{51} & \ding{55} & \ding{55} & \ding{55} & \ding{51} \\
\midrule
\benchmark{} (ours)                      & \ding{51} & \ding{51} & \ding{51} & \ding{51} & \ding{51} \\
\bottomrule
\end{tabular}
\vspace{-5pt}
\end{table}
\section{Failure Analysis: The Limits of Prompting and Explicit Reasoning}
\label{sec:behavioral}

\begin{table*}[t]
\caption{Accuracy cost per saved call ($\frac{\Delta\text{Acc}}{-\Delta\text{TC}}$) when switching from Default~($\star$) to Sparse~(S). All changes are relative to Default (Prompt-only). More negative $\frac{\Delta\text{Acc}}{-\Delta\text{TC}}$ means each saved call costs more accuracy. Hard tasks pay a disproportionate price.}
\vspace{-5pt}
\label{tab:uniform_reduction}
\centering
\small
\setlength{\tabcolsep}{3pt}
\begin{tabular}{ll rrr rrr rrr}
\toprule
& & \multicolumn{3}{c}{Qwen3-4B-Inst.} & \multicolumn{3}{c}{Qwen3-14B} & \multicolumn{3}{c}{Llama-3.3-70B} \\
\cmidrule(lr){3-5} \cmidrule(lr){6-8} \cmidrule(lr){9-11}
Mode & Difficulty & $\Delta$Acc & $\Delta$TC & $\frac{\Delta\text{Acc}}{-\Delta\text{TC}}$ & $\Delta$Acc & $\Delta$TC & $\frac{\Delta\text{Acc}}{-\Delta\text{TC}}$ & $\Delta$Acc & $\Delta$TC & $\frac{\Delta\text{Acc}}{-\Delta\text{TC}}$ \\
\midrule
\multirow{3}{*}{Prompt-only}
  & Easy  & $-$14.5 & $-$0.84 & $-$17.3 & $-$8.8  & $-$0.59 & $-$14.9 & $+$1.6  & $-$0.51 & $+$3.2 \\
  & Medium& $-$20.7 & $-$0.86 & $-$24.1 & $-$12.9 & $-$0.53 & $-$24.3 & $+$2.0  & $-$0.41 & $+$4.8 \\
  & Hard  & $-$20.3 & $-$0.48 & $-$\textbf{42.4} & $-$27.3 & $-$0.47 & $-$\textbf{58.4} & $-$0.2  & $-$0.34 & $-$\textbf{0.5} \\
\midrule
\multirow{3}{*}{Reason-then-Act}
  & Easy  & $-$14.5 & $-$0.86 & $-$16.9 & $-$4.4  & $-$0.67 & $-$6.6  & $-$4.8  & $-$1.98 & $-$2.4 \\
  & Medium& $-$22.4 & $-$0.90 & $-$24.8 & $-$10.4 & $-$0.62 & $-$16.8 & $-$18.9 & $-$1.87 & $-$10.1 \\
  & Hard  & $-$13.0 & $-$0.35 & $-$\textbf{36.6} & $-$9.7  & $-$0.28 & $-$\textbf{34.7} & $-$63.3 & $-$1.99 & $-$\textbf{31.7} \\
\bottomrule
\end{tabular}
\end{table*}

\begin{figure}[t]
\centering
\includegraphics[width=\textwidth]{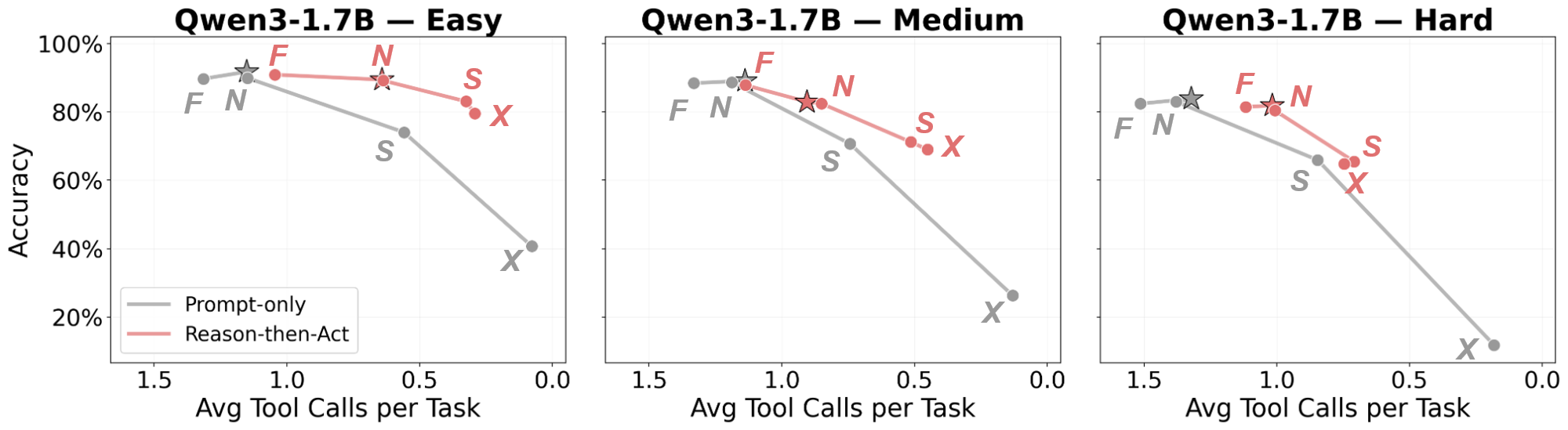}
\vspace{-15pt}
\caption{Accuracy vs.\ Avg tool calls per difficulty for Qwen3-1.7B. $\star$=Default, F=Force, N=Necessary, S=Sparse, X=No~Tool. Reason-then-Act (red) partially shifts the tradeoff, reducing unnecessary easy-task calls, but still produces negative efficiency on hard tasks. Lines fold non-monotonically, reflecting coarse prompt control.}
\vspace{-10pt}
\label{fig:baselines}
\end{figure}

With \benchmark{} in place, we systematically test whether models can calibrate their tool usage through the two most natural training-free approaches: varying the prompt (Prompt-only) and asking the model to explicitly reason about tool necessity before acting (Reason-then-Act). Surprisingly, we found that these approaches based on prompt engineering and reasoning \textit{fail} to selectively reduce unnecessary tool calls as we show in the experiments. 

\subsection{Experimental setup}

We evaluate six models spanning two families: Qwen3-1.7B/4B/14B/32B and Llama-3.1-8B/3.3-70B. All experiments are run 3 times with different random seeds and we report mean results.

\paragraph{Prompt-only baselines.}
We test five prompt modes spanning the full range of tool-use instructions:
\textbf{Force~(F)}, ``tool use is mandatory'';
\textbf{Default~($\star$)}, no explicit requirements;
\textbf{Necessary~(N)}, ``only if necessary'';
\textbf{Sparse~(S)}, ``expensive, use sparingly'';
and \textbf{No~Tool~(X)}, ``do not use any tools.''

\paragraph{Reason-then-act baselines.}
In addition to prompt-only control, we evaluate a stronger baseline inspired by the think-before-act paradigm from ReAct~\citep{yao2022react} and Reflexion~\citep{shinn2023reflexion}. Before making a tool-call decision, the model is instructed to first reason about whether it can solve the task directly or needs a tool, then act on its own assessment. We apply this reasoning step to each of the same five prompt modes (Force, Default, Necessary, Sparse, No~Tool).

\subsection{Key findings}

\paragraph{Finding 1: Models default to tool overuse.}
Under the Default~($\star$) setting in Prompt-only baselines, models make 2,100--4,400 total tool calls across the 2,250-task single-hop test set, more than one calls per task. Even on easy tasks, Qwen3-1.7B makes 864 tool calls out of 750 easy tasks, and Llama-3.3-70B makes 1,482. The model's default behavior is ``tools are available, therefore use them,'' even when the task is simple enough to solve directly.
\vspace{-5pt}
\paragraph{Finding 2: Prompt engineering reduces tool calls indiscriminately, and hard tasks pay a disproportionate price.}
Prompts that discourage tool usage reduce calls across \emph{all} difficulty levels, including hard tasks where tools are genuinely needed. The reduction is far too indiscriminate: hard tasks lose substantially more accuracy per saved call than easy tasks. We quantify this with the \emph{accuracy cost per saved call}, $\frac{\Delta\text{Acc}}{-\Delta\text{TC}}$: how much accuracy is lost for each tool call eliminated. More negative means each saved call is more costly. Table~\ref{tab:uniform_reduction} shows this cost when moving from Default~($\star$) to Sparse~(S). On Qwen3-4B-Instruct, the cost is $-$17.3 on easy but reaches $-$42.4 on hard, meaning hard tasks lose 2.5$\times$ more accuracy per saved call. This pattern holds across model sizes.
\vspace{-5pt}
\paragraph{Finding 3: Reason-then-Act only partially helps, but with additional cost.}
Reasoning before decisions partially mitigates this problem: the accuracy cost per saved call improves on easy tasks (e.g., Qwen3-14B easy improves from $-$14.9 to $-$6.6 in Table~\ref{tab:uniform_reduction}). Figure~\ref{fig:baselines} shows Reason-then-Act (red lines) sits closer to the upper-right region on easy tasks, indicating that reasoning does reduce some unnecessary calls. However, reasoning still carries a high accuracy cost per saved call on hard tasks (e.g., $-$34.7 for Qwen3-14B), since it also suppresses tool calls where they are genuinely needed. Critically, this partial improvement comes at a cost: reasoning requires the model to generate additional tokens, increasing the generation overhead. Moreover, reasoning is model-dependent: on Llama-3.1-8B, accuracy drops from 79.5\% to 31.2\%; on Llama-3.3-70B, from 83.1\% to 47.9\%, as the model narrates its intent to call tools but never produces a valid invocation, resulting in near-zero tool calls (see Table~\ref{tab:singlehop_full} in Appendix~\ref{app:full_reduction} for full details).
\vspace{-5pt}
\paragraph{Finding 4: Prompt engineering cannot precisely control the accuracy--tool-call tradeoff.}
In practice, a user may want to set a tool-call budget and maximize accuracy under that budget. Figure~\ref{fig:baselines} shows that neither Prompt-Only nor Reason-then-Act can achieve this: each prompt mode provides a single, fixed operating point, with no way to smoothly adjust the tradeoff. Furthermore, many of these operating points are nearly indistinguishable: instructing the model to use tools ``only if necessary'' (N) produces almost identical behavior to the neutral Default~($\star$), while in Reason-then-Act mode, Sparse~(S) and No~Tool~(X) collapse to nearly the same point. The two baselines offer only a few effective operating points, making it impossible to precisely target a desired budget.

These findings establish that prompt-level control over tool-call decisions is limited, coarse, and unreliable across models. The models appear to lack the ability to translate task understanding into calibrated tool-call decisions through the generation process. This raises the question: is the model \emph{unable} to assess tool necessity, or does it know internally but fail to act on it? We investigate this in the next section.

\section{Probing Analysis: Decoding Implicit Tool Necessity}
\label{sec:probing}

\begin{table}[b!]
\vspace{-10pt}
\caption{AUROC and accuracy for predicting tool necessity from pre-generation hidden states. All probes achieve high AUROC, confirming that tool necessity is consistently encoded in models.}
\label{tab:probe_results}
\small
\centering
\begin{tabular}{lcccccc}
\toprule
& & & \multicolumn{3}{c}{AUROC by difficulty} \\
\cmidrule(lr){4-6}
Model & AUROC & Acc & Easy & Med & Hard \\
\midrule
Qwen3-1.7B            & 0.894 & 0.847 & 0.864 & 0.831 & 0.904 \\
Qwen3-4B-Inst.        & 0.948 & 0.877 & 0.933 & 0.906 & 0.948 \\
Llama-3.1-8B-Inst.    & 0.927 & 0.849 & 0.892 & 0.867 & 0.884 \\
Qwen3-14B             & 0.957 & 0.892 & 0.955 & 0.907 & 0.941 \\
Qwen3-32B             & 0.952 & 0.885 & 0.951 & 0.903 & 0.939 \\
Llama-3.3-70B-Inst.   & 0.936 & 0.872 & 0.906 & 0.849 & 0.956 \\
\bottomrule
\end{tabular}
\end{table}

We now investigate whether models encode tool-necessity information internally, even when they fail to act on it during generation. Interestingly, we find that \textit{tool necessity is already encoded in hidden states} as shown in the experiments.

\paragraph{Setup.}
For each task, we first collect a binary label: we force the model to answer without tool access, and label tasks where it succeeds as tool-unnecessary ($y{=}0$) and tasks where it fails as tool-necessary ($y{=}1$). We then extract hidden states by running a single forward pass and taking the hidden state at the last token position across all layers. Finally, we concatenate all-layer features and train an L2-regularized logistic regression to predict tool necessity. The probe trains on 900 training examples and is evaluated on 2,250 held-out test tasks. Training takes seconds on CPU.
\vspace{-5pt}
\paragraph{Tool necessity is linearly decodable.}
The probe achieves AUROC 0.89--0.96 across all six tested models (Table~\ref{tab:probe_results}), confirming that tool necessity is consistently encoded in pre-generation hidden states regardless of model family or size. Even the smallest model carries a strong signal, while larger models reach 0.95+. The per-difficulty breakdown shows strong performance across all levels, with medium tasks being the most challenging, consistent with medium being the decision boundary where the model is uncertain.
\vspace{-5pt}
\paragraph{The signal exists even when generation fails.}
The most striking evidence comes from the Llama models. As discussed in Section~\ref{sec:behavioral}, Reason-then-Act completely breaks tool calling on Llama-3.1-8B (79.5\% $\to$ 31.2\%) and Llama-3.3-70B (83.1\% $\to$ 47.9\%). Yet the linear probe still achieves AUROC above 0.9; the information about tool necessity is clearly present in the representation, even though the model is entirely unable to express it during generation. 

Based on these findings, our next step is to use this signal to directly steer the model's tool-call behavior. In the next section, we show how injecting a short steering sentence into the model's output can translate this hidden knowledge into better tool-call decisions.

\section{\method{}: Turning Hidden Knowledge into Better Decisions}
\label{sec:method}

In this section, we propose \method{}, a lightweight inference-time method that uses the probe's prediction to prefill the model's response with a short steering sentence, guiding the model to either solve directly or call a tool. This translates the internal signal identified in Section~\ref{sec:probing} into better tool-call decisions, requiring no model fine-tuning and no reasoning overhead.

\subsection{Method}

\method{} operates in three steps at inference time:

\paragraph{Step 1: Extract last hidden states.}
Given a task prompt, we run a single forward pass over the input tokens and extract the hidden states at the last token position. This is the standard prompt-encoding step that every LLM performs to get KV cache before autoregressive decoding begins, so hidden state extraction adds no additional forward passes.
\vspace{-5pt}
\paragraph{Step 2: Linear probe prediction.}
We apply the trained linear probe to all-layer hidden states, producing a probability $p$. A threshold $\tau$ converts this into a binary decision: if $p < \tau$, the task is predicted to be solvable without tools; otherwise, a tool call is predicted to be necessary. The threshold $\tau$ provides a single knob for controlling the accuracy--efficiency tradeoff: lower $\tau$ skips more tool calls (saves cost, risks missing necessary ones); higher $\tau$ preserves more tool calls (higher accuracy, fewer savings).
\vspace{-5pt}
\paragraph{Step 3: Prefill before generation.}
Based on the probe's prediction, we prepend a short steering sentence to the beginning of the model's response:
\begin{itemize}[leftmargin=*]
    \vspace{-3pt}
    \item If $p < \tau$ (tool unnecessary): ``\texttt{I can solve this directly without using a tool.}''
    \vspace{-3pt}
    \item If $p \geq \tau$ (tool necessary): ``\texttt{I need to use a tool for this question.}''
    \vspace{-3pt}
\end{itemize}
The model then continues generating from this prefill, producing either a direct answer or a tool call. This \emph{soft} prefill allows the model to override the suggestion if its own assessment disagrees. We also evaluate a \emph{hard} prefill mode that forces the output format (\texttt{\textbackslash boxed\{} for direct answers, tool-call JSON for tool use), leaving no room for the model to deviate (Table~\ref{tab:abl_prefill}, Appendix~\ref{app:ablation_prefill}).

\subsection{Main results}

\begin{figure*}[t]
\centering
\includegraphics[width=\textwidth]{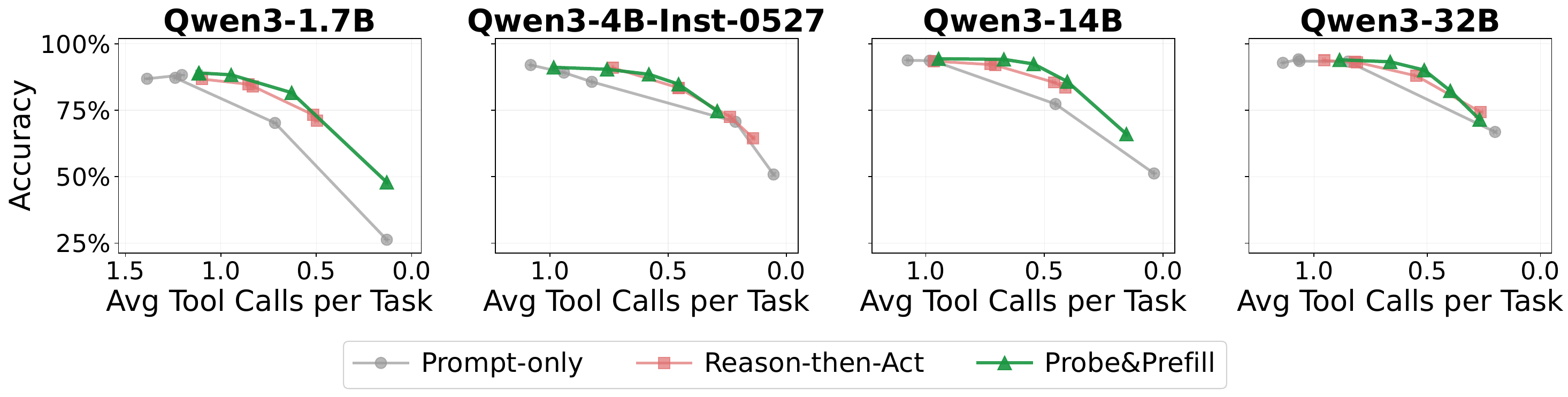}
\vspace{-15pt}
\caption{Accuracy vs.\ total tool calls for Qwen models. \method{} sweeps threshold $\tau$ from 0.1 to 0.9, achieving a strictly better tradeoff than both baselines.}
\vspace{-5pt}
\label{fig:tradeoff}
\end{figure*}

\begin{figure*}[t]
\centering
\includegraphics[width=0.8\textwidth]{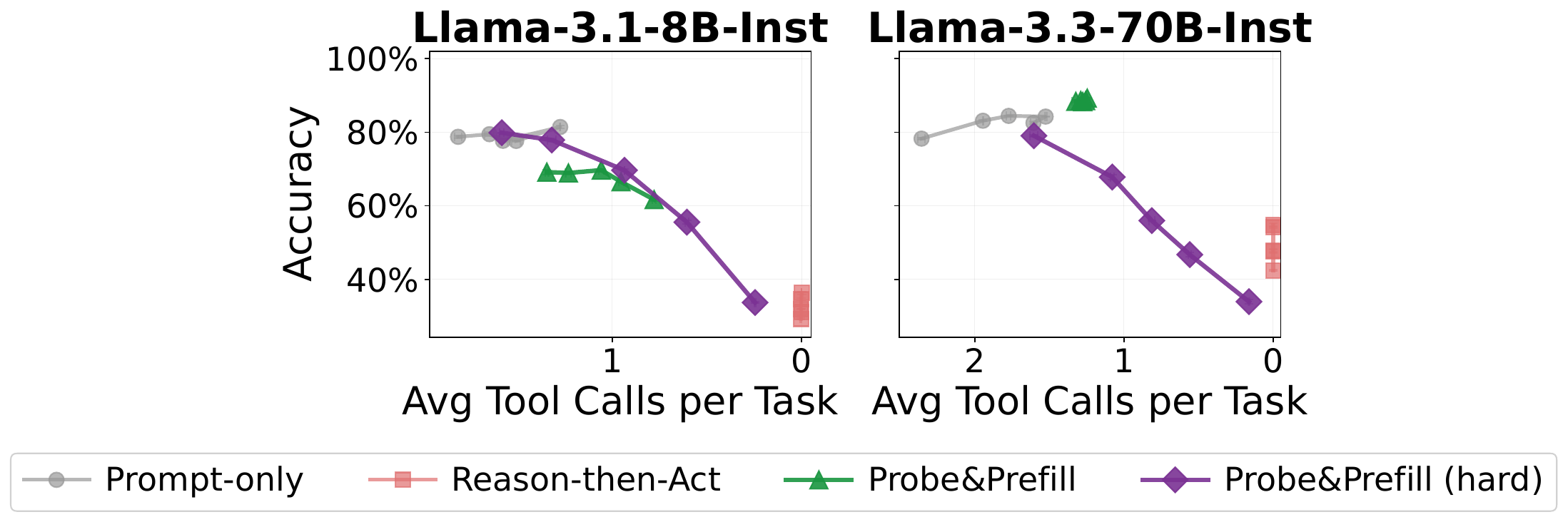}
\vspace{-5pt}
\caption{Accuracy vs.\ total tool calls for Llama models. Soft prefill (green) is partially ignored; hard prefill (purple) forces the output format and restores full tradeoff control.}
\vspace{-10pt}
\label{fig:tradeoff_llama}
\end{figure*}

Figures~\ref{fig:tradeoff} and~\ref{fig:tradeoff_llama} show the accuracy vs.\ tool-call tradeoff across six models.

\paragraph{\method{} outperforms the Prompt-only baseline.}
Compared to the Prompt-only baseline (gray), \method{} (green) is strictly better on Qwen models (Figure~\ref{fig:tradeoff}): it achieves higher accuracy at every tool-call budget, or equivalently, fewer tool calls at every accuracy level. Moreover, by sweeping the threshold $\tau$, \method{} traces a smooth tradeoff curve, providing fine-grained control over the operating point. In contrast, prompt engineering offers only a handful of discrete, fixed operating points with no way to interpolate between them. For Llama models (Figure~\ref{fig:tradeoff_llama}), soft prefill is partially ignored due to the bad instruction following; hard prefill restores full tradeoff control by forcing the output format.

\vspace{-5pt}
\paragraph{\method{} outperforms Reason-then-Act.}
\method{} also outperforms the Reason-then-Act baseline (red) in most cases, despite requiring no additional reasoning tokens. Reasoning asks the model to verbally assess tool necessity before acting, yet the probe's prediction from the pre-generation hidden state is more accurate than the model's own verbalized assessment. This suggests that reasoning about tool necessity is largely superficial: it does not improve the model's underlying decision beyond what is already encoded in its hidden states before any reasoning takes place. A striking example comes from the Llama models, where reasoning completely collapses tool calling (the red line in Figure~\ref{fig:tradeoff_llama} drops to near-zero tool calls with large accuracy loss), yet \method{} still achieves strong performance by reading the hidden state directly.
\vspace{-5pt}
\paragraph{Adaptive tool-call reduction.}
A key advantage of \method{} over the baselines is \emph{adaptive} reduction: the probe selectively skips easy calls while preserving hard ones. Table~\ref{tab:adaptive_reduction} compares the accuracy cost per saved call across all prompt strategies and \method{} ($\tau$=0.5), averaged over six models. Every baseline shows strongly negative costs, meaning each saved call comes at a significant accuracy penalty, especially on hard tasks. \method{} achieves the lowest cost across all difficulty levels ($-$1.6 easy, $-$3.4 hard), reducing tool calls with minimal accuracy loss. 

\begin{table*}[h]
\caption{Accuracy cost per saved call ($\frac{\Delta\text{Acc}}{-\Delta\text{TC}}$), averaged across six models (threshold $\tau$ is set to 0.5 for \method{}). All changes are relative to Default~($\star$, Prompt-only). More negative $\frac{\Delta\text{Acc}}{-\Delta\text{TC}}$ means each saved call costs more accuracy. \method{} achieves lowest cost per saved call.}
\vspace{-5pt}
\label{tab:adaptive_reduction}
\centering
\footnotesize
\setlength{\tabcolsep}{1.8pt}
\begin{tabular}{l rrr rrr rrr rrr}
\toprule
& \multicolumn{3}{c}{Easy} & \multicolumn{3}{c}{Medium} & \multicolumn{3}{c}{Hard} & \multicolumn{3}{c}{Overall} \\
\cmidrule(lr){2-4} \cmidrule(lr){5-7} \cmidrule(lr){8-10} \cmidrule(lr){11-13}
Method & $\Delta$Acc & $\Delta$TC & $\frac{\Delta\text{Acc}}{-\Delta\text{TC}}$ & $\Delta$Acc & $\Delta$TC & $\frac{\Delta\text{Acc}}{-\Delta\text{TC}}$ & $\Delta$Acc & $\Delta$TC & $\frac{\Delta\text{Acc}}{-\Delta\text{TC}}$ & $\Delta$Acc & $\Delta$TC & $\frac{\Delta\text{Acc}}{-\Delta\text{TC}}$ \\
\midrule
Necessary (N)     & $-$0.3  & $-$0.09 & $-$3.5  & $-$0.7  & $-$0.04 & $-$17.9 & $-$1.9  & $-$0.04 & $-$44.4 & $-$1.0  & $-$0.06 & $-$16.8 \\
Necessary + Reason-then-Act & $-$8.1  & $-$0.95 & $-$8.5  & $-$16.3 & $-$0.82 & $-$19.7 & $-$23.2 & $-$0.70 & $-$32.9 & $-$15.8 & $-$0.82 & $-$19.2 \\
Sparse (S)        & $-$6.3  & $-$0.55 & $-$11.3 & $-$7.9  & $-$0.47 & $-$16.9 & $-$11.1 & $-$0.35 & $-$31.5 & $-$8.4  & $-$0.46 & $-$18.4 \\
Sparse + Reason-then-Act    & $-$9.9  & $-$1.13 & $-$8.8  & $-$19.9 & $-$1.04 & $-$19.1 & $-$29.3 & $-$0.84 & $-$34.7 & $-$19.7 & $-$1.00 & $-$19.6 \\
No Tool (X)       & $-$18.1 & $-$0.72 & $-$25.2 & $-$26.7 & $-$0.72 & $-$36.9 & $-$41.4 & $-$0.69 & $-$60.4 & $-$28.7 & $-$0.71 & $-$40.5 \\
No Tool + Reason-then-Act   & $-$12.4 & $-$1.18 & $-$10.5 & $-$23.5 & $-$1.12 & $-$20.9 & $-$36.6 & $-$0.94 & $-$39.1 & $-$24.2 & $-$1.08 & $-$22.4 \\
\midrule
\method{}  (Ours)        & $-$1.1  & $-$0.66 & \textbf{$-$1.6}  & $-$3.4  & $-$0.54 & \textbf{$-$6.2}  & $-$0.8  & $-$0.24 & \textbf{$-$3.4}  & $-$1.7  & $-$0.48 & \textbf{$-$3.6} \\
\bottomrule
\end{tabular}
\end{table*}

\subsection{Additional experiments}

Since \method{} introduces a learned probe, it is important to verify that results are not artifacts of overfitting to the training environments or sensitive to hyperparameter choices. We conduct the following additional experiments (full details in the Appendix):
\begin{itemize}[leftmargin=*]
    \item \textbf{Complete per-model results} (Appendix~\ref{app:full_reduction}): Full numerical tables for all models, prompt modes, reasoning settings, and probe thresholds on single-hop tasks.
    \item \textbf{Multi-hop tasks} (Appendix~\ref{app:multihop}): On the 3 multi-hop environments, \method{} reduces tool calls by up to 75\% on Qwen models while maintaining or improving accuracy.
    \item \textbf{Out-of-distribution transfer} (Appendix~\ref{app:ood}): The probe trained on a subset of environments generalizes to held-out environments within the same category.
    \item \textbf{Ablation studies} (Appendix~\ref{app:ablations}): Soft vs.\ hard prefill, temperature scaling, layer selection, data efficiency, and regularization strength.
    \item \textbf{Inference overhead} (Appendix~\ref{app:overhead}): The probe adds $<$1ms per task on top of the standard prefill forward pass, with no additional model calls.
    \item \textbf{Real-world agentic search} (Appendix~\ref{app:search_o1}): On Search-o1 agentic benchmark, \method{} reduces search API calls by up to 56\% while matching or exceeding baseline accuracy.
    \item \textbf{Supervised training baselines} (Appendix~\ref{app:sft}): We compare with full SFT on 900 examples and with full SFT and rank-16 LoRA on 6{,}000 examples; these training recipes do not consistently reduce tool calls.
    \item \textbf{Confidence-based controllers} (Appendix~\ref{app:confidence_controllers}): Verbalized confidence, first-token likelihood, and self-ask drive the same soft-prefill mechanism.
    \item \textbf{Difficulty supervision control} (Appendix~\ref{app:difficulty_supervision}): We compare necessity labels with an easy-versus-hard training signal.
\end{itemize}
\section{Related Work}
\label{sec:related}

\paragraph{Agentic Tool-use benchmarks.}
Several benchmarks evaluate LLM tool use. ToolQA~\citep{zhuang2023toolqa} tests question answering over external data sources; API-Bank~\citep{li2023api} evaluates API selection across hundreds of real APIs; Toolformer~\citep{schick2023toolformer} and ToolLLM~\citep{qin2023toolllm} train models to invoke tools correctly; Gorilla~\citep{patil2024gorilla} targets correct API call generation; and BFCL~\citep{patil2025berkeley} provides a comprehensive function-calling leaderboard spanning single-turn, multi-turn, and agentic settings. All these benchmarks evaluate \emph{whether} models can use tools correctly, assuming every task requires a tool. In contrast, \benchmark{} evaluates the tool-call \emph{decision}: given the correct tool, the model must decide whether to use it or solve directly (Table~\ref{tab:benchmark_compare}).

\paragraph{Efficient tool calling.}
Recent work on reducing unnecessary tool calls takes several approaches. \citet{xu2025alignment} fine-tune the model to invoke tools only when its confidence is low, reducing calls by $\sim$50\% on arithmetic and QA tasks. \citet{wu2025joint} jointly optimize agent instructions and tool descriptions via verbalized feedback, reducing calls by up to 70\%. \citet{yang2026toward} survey efficiency across memory, tool learning, and planning in LLM agents. These works directly build costly interventions, such as SFT pipelines or iterative prompt optimization, without first investigating \emph{why} models overcall. We instead take a mechanistic approach in a more realistic agentic setting, showing that the model's hidden state already encodes the tool-call decision and can be leveraged with only a lightweight linear probe that takes seconds to train and adds negligible inference cost.

\paragraph{Probing and controlling LLM behavior.}
Linear probing has revealed that LLM hidden states encode syntactic structure~\citep{hewitt2019structural}, factual knowledge~\citep{burns2022discovering}, truthfulness~\citep{li2023inference}, and self-knowledge~\citep{kadavath2022language}. Building on these findings, a growing body of work uses internal representations to steer model behavior: activation addition~\citep{turner2024activation} and representation engineering~\citep{zou2023representation} modify hidden activations during generation, concept bottleneck LLMs~\citep{sun2024crafting,sun2024concept} route predictions through interpretable concept layers, and recent work edits model weights guided by steering directions for reasoning control~\citep{sun2025thinkedit, sun2026steer2edit, yan2025reflctrl} and skill unlearning~\citep{li2025effective}. Our work extends probing to a new domain, agentic tool-use decisions, showing that models ``mostly know'' when they need tools but fail to act on it. Unlike prior steering methods that modify activations or weights, we steer via output prefilling based on a probe prediction, requiring no modification to the forward pass and remaining compatible with any serving infrastructure.

\section{Conclusion}
\label{sec:conclusion}

In this work, we showed that LLM agents already encode reliable tool-necessity signals in their hidden states, even when they fail to act on them during generation. A simple linear probe extracts this signal, and prefilling the model's response based on the probe prediction yields a strictly better accuracy--efficiency tradeoff than both Prompt-only and Reason-then-Act baselines. Our benchmark, probing analysis, and method together demonstrate that lightweight, training-free interventions can meaningfully improve agent tool-use efficiency.

\section*{Acknowledgement}
C. Sun, G. Yan, and T.-W. Weng are partially supported by National Science Foundation under Grant No. 2313105, 2430539, and Intel Rising Star Faculty Award. The authors would like to thank the National Research Platform (NRP) at UCSD for computational support and anonymous reviewers for valuable feedback.

\bibliographystyle{plainnat}
\bibliography{references}

\newpage
\appendix

\addcontentsline{toc}{section}{Appendix}
\part{}
\parttoc
\newpage

\section{Benchmark environment details}
\label{app:envs}

Table~\ref{tab:env_details} provides an overview of all 15 environments. Below we describe each environment in detail, including its real-world motivation, available tools, answer format, and how difficulty levels are constructed.

\begin{table}[H]
\caption{Overview of \benchmark{} environments. Single-hop envs have 20 train / 50 test tasks per difficulty. Multi-hop envs follow the same split with 3-hop chained tasks ($x \to y \to z$).}
\label{tab:env_details}
\centering
\small
\begin{tabular}{lllp{6.5cm}}
\toprule
Category & Type & Environment & Example (easy $\to$ hard) \\
\midrule
\multirow{6}{*}{A: Scale}
  & \multirow{5}{*}{1-hop} & CalculatorEnv & ``$20+20$'' $\to$ ``$(10^{12} \times 10^{11}) - 10^8$'' \\
  & & StatisticsEnv & ``Mean of [5,7,13]'' $\to$ ``Correlation of 25-element lists'' \\
  & & CountingEnv & ``$C(6,4)$'' $\to$ ``$C(80,40)$'' \\
  & & MatrixEnv & ``det of 2$\times$2'' $\to$ ``det of 5$\times$5'' \\
  & & PrimeEnv & ``Is 29 prime?'' $\to$ ``Is 104729 prime?'' \\
  \cmidrule(lr){2-4}
  & 3-hop & ChainedCalculatorEnv & $x{=}40{-}10,\ y{=}x{+}5,\ z{=}y{-}19$ $\to$ trillion-scale chains \\
\midrule
\multirow{6}{*}{B: Knowledge}
  & \multirow{5}{*}{1-hop} & RetrieverEnv & ``Capital of France'' $\to$ ``Synthetic entity lookup'' \\
  & & HistoricalYearEnv & ``Moon landing year'' $\to$ ``Fictional event year'' \\
  & & GameRuleEnv & ``Chess pieces per player'' $\to$ ``Fictional game stats'' \\
  & & HashEnv & ``MD5 of `hello'\,'' $\to$ ``Custom hash algorithm'' \\
  & & DecodingEnv & ``Morse: SOS'' $\to$ ``Custom cipher decode'' \\
  \cmidrule(lr){2-4}
  & 3-hop & ChainedRetrieverEnv & Look up $x$, use $x$ to find $y$, look up $z$ about $y$ \\
\midrule
\multirow{6}{*}{C: Execution}
  & \multirow{5}{*}{1-hop} & ListManipulationEnv & ``Remove from [1,2,3]'' $\to$ ``2D list operations'' \\
  & & DateTimeEnv & ``Days in April'' $\to$ ``Day-of-week for arbitrary date'' \\
  & & CodeExecutorEnv & ``print(2+3)'' $\to$ ``Trace 20-iteration DP'' \\
  & & ScheduleEnv & ``2 meetings overlap?'' $\to$ ``10+ meetings, find free slots'' \\
  & & RegexMatchEnv & ``\textbackslash d+ on `abc123'\,'' $\to$ ``Complex regex on 60-char string'' \\
  \cmidrule(lr){2-4}
  & 3-hop & ChainedCodeExecutorEnv & Run code$_1 \to x$, code$_2(x) \to y$, code$_3(y) \to z$ \\
\bottomrule
\end{tabular}
\end{table}

\subsection{Design principles}

\benchmark{} is designed to be lightweight, zero-cost, and easily extensible. All environments run locally with no API keys, external services, or network access required. New environments or difficulty levels can be added by writing a task generator with a fixed random seed.

\begin{enumerate}
    \item \textbf{Zero-cost, fully offline}: All tool responses are simulated locally and deterministically. No paid APIs, no network calls, no rate limits. The entire benchmark can be run on a single machine.
    \item \textbf{Exact-form answers}: Answers are numbers, strings, or lists that can be verified programmatically with no ambiguity or need for LLM-based judging.
    \item \textbf{Short inputs}: Task descriptions are 1--15 lines, keeping prompt overhead low and ensuring the tool-call decision is the primary challenge.
    \item \textbf{Unambiguous category membership}: Each environment belongs to exactly one category, enabling clean ablation across tool-necessity types.
    \item \textbf{Deterministic and reproducible}: All tasks are generated with fixed random seeds. Re-running any generator produces byte-identical output.
    \item \textbf{Easily extensible}: Adding a new environment requires only a self-contained task generator script. The benchmark grew from 15 single-hop to 18 environments (including 3 multi-hop) with no changes to the evaluation infrastructure.
\end{enumerate}

\subsection{Evaluation framework}

\benchmark{} reports two quantities per model per setting:

\paragraph{Accuracy.}
Each task has an exact-form expected answer (number, string, or list). The model's final response is extracted from the \texttt{\textbackslash boxed\{\}} output format and compared against the ground truth using a deterministic evaluator that handles numeric tolerance, case-insensitive string matching, and equivalent representations (e.g., different date formats). No LLM-based judging is used.

\paragraph{Total tool calls.}
We count the total number of tool calls made across the test set. Combined with the three difficulty levels (easy, medium, hard), this allows users to compare how different models or methods trade off accuracy against tool usage at each difficulty, revealing whether a method reduces calls indiscriminately or adaptively targets unnecessary ones.

\paragraph{Reproducibility.}
All tasks are generated with fixed random seeds. All experiments are run 3 times with different random seeds; we report mean and standard deviation. The benchmark, evaluation code, and all task generators will be released upon acceptance.

\subsection{Category A: Computational scale}

These environments test whether the model can assess the limits of its own mental arithmetic. The model understands the operation in every case; the question is whether the numbers or data involved exceed what it can compute reliably.

\subsubsection{CalculatorEnv}

\textbf{Motivation.}
Arithmetic is the most basic tool-use scenario for any LLM agent. Agents that assist with financial calculations, scientific computations, or everyday math must decide whether an expression is simple enough to compute directly or requires calling a calculator. This environment isolates that decision by varying only the magnitude of the operands.

\textbf{Tools.}
\begin{itemize}
    \item \texttt{evaluate\_expression(expr)}: Evaluates a mathematical expression string and returns the exact numerical result.
    \item \texttt{get\_last\_result()}: Returns the result of the most recent evaluation, useful for multi-step calculations.
    \item \texttt{clear\_last\_result()}: Clears the stored result.
\end{itemize}

\textbf{Answer format.} Exact number.

\textbf{Difficulty levels.}
\begin{itemize}
    \item \textbf{Easy}: Small numbers (2--40) with 5 expression templates: \texttt{a+b}, \texttt{a-b}, \texttt{a*b}, \texttt{(a+b)-c}, \texttt{(a*b)+c}.\\
    \emph{Example}: ``Compute exactly: $20 + 20$'' $\to$ \textbf{40}
    \item \textbf{Medium}: 3-digit numbers (80--900) with multiplication, division, and modulo.\\
    \emph{Example}: ``Compute exactly: $(810 \times 87) - 85 + 178$'' $\to$ \textbf{70563}
    \item \textbf{Hard}: Numbers in $10^9$--$10^{12}$, far exceeding mental computation.\\
    \emph{Example}: ``Compute exactly: $(39006255142 \times 342002902703) - 702386298$'' $\to$ \textbf{13340252482137117062528}
\end{itemize}

\subsubsection{StatisticsEnv}

\textbf{Motivation.}
Data analysis agents frequently need to compute summary statistics, such as means, standard deviations, and correlations, to generate reports, evaluate A/B tests, or summarize datasets. While simple averages of a few numbers are easy to compute mentally, statistics involving larger datasets or more complex measures (e.g., Pearson correlation) quickly become infeasible without a tool.

\textbf{Tools.}
\begin{itemize}
    \item \texttt{compute\_stat(data, stat\_type)}: Computes a specified statistic (\texttt{mean}, \texttt{median}, \texttt{std}, \texttt{percentile}, \texttt{correlation}, etc.) on the provided data and returns the exact result.
    \item \texttt{describe(data)}: Returns a full descriptive summary (count, mean, std, min, quartiles, max) for the dataset.
\end{itemize}

\textbf{Answer format.} Exact number to specified precision.

\textbf{Difficulty levels.}
\begin{itemize}
    \item \textbf{Easy}: Mean or median of 3--5 small numbers.\\
    \emph{Example}: ``What is the median of [3, 7, 1, 9, 5]?'' $\to$ \textbf{5}
    \item \textbf{Medium}: Standard deviation or percentiles of 8--15 numbers.\\
    \emph{Example}: ``What is the standard deviation of [12, 15, 18, 22, 25, 30, 14, 19, 27, 11]? Round to 2 decimal places.'' $\to$ \textbf{6.33}
    \item \textbf{Hard}: Pearson correlation on 20--30 numbers.\\
    \emph{Example}: ``What is the Pearson correlation between X=[...20 numbers...] and Y=[...20 numbers...]? Round to 4 decimal places.'' $\to$ \textbf{0.9994}
\end{itemize}

\subsubsection{CountingEnv}

\textbf{Motivation.}
Combinatorial calculations arise in planning, scheduling, and resource allocation tasks. An agent planning team assignments or seating arrangements may need to compute combinations or permutations. Small values are manageable mentally, but combinatorial results grow extremely fast, making tools essential for larger inputs.

\textbf{Tools.}
\begin{itemize}
    \item \texttt{combination(n, k)}: Computes $\binom{n}{k}$ and returns the exact integer.
    \item \texttt{permutation(n, k)}: Computes $P(n,k) = \frac{n!}{(n-k)!}$ and returns the exact integer.
    \item \texttt{factorial(n)}: Computes $n!$ and returns the exact integer.
\end{itemize}

\textbf{Answer format.} Exact integer.

\textbf{Difficulty levels.}
\begin{itemize}
    \item \textbf{Easy}: Small factorials and combinations ($C(8,3)$, $6!$).\\
    \emph{Example}: ``How many ways can you choose 2 items from 5?'' $\to$ \textbf{10}
    \item \textbf{Medium}: Larger combinations ($C(20,7)$, $14!$).\\
    \emph{Example}: ``Compute $P(15,4)$.'' $\to$ \textbf{32760}
    \item \textbf{Hard}: Very large values ($C(80,30)$, $25!$).\\
    \emph{Example}: ``What is $C(50,25)$?'' $\to$ \textbf{126410606437752}
\end{itemize}

\subsubsection{MatrixEnv}

\textbf{Motivation.}
Matrix operations are fundamental to ML engineering (weight matrices, attention computations), computer graphics (transformations), and scientific computing (solving linear systems). Computing a $2\times2$ determinant is straightforward, but determinants of larger matrices involve recursive expansion with many terms, making mental computation error-prone.

\textbf{Tools.}
\begin{itemize}
    \item \texttt{matrix\_determinant(matrix)}: Computes the determinant of a square matrix and returns the exact value.
    \item \texttt{matrix\_multiply(A, B)}: Computes the product of two matrices and returns the result matrix.
    \item \texttt{matrix\_trace(matrix)}: Computes the trace (sum of diagonal elements) and returns the exact value.
\end{itemize}

\textbf{Answer format.} Exact number or matrix.

\textbf{Difficulty levels.}
\begin{itemize}
    \item \textbf{Easy}: $2\times2$ determinant or trace.\\
    \emph{Example}: ``What is the trace of $\begin{bmatrix}3&1\\7&4\end{bmatrix}$?'' $\to$ \textbf{7}
    \item \textbf{Medium}: $3\times3$ determinant.\\
    \emph{Example}: ``What is the determinant of $\begin{bmatrix}2&3&1\\4&1&3\\1&2&4\end{bmatrix}$?'' $\to$ \textbf{$-$17}
    \item \textbf{Hard}: $4\times4$ or $5\times5$ determinant, where cofactor expansion requires tracking dozens of terms.
\end{itemize}

\subsubsection{PrimeEnv}

\textbf{Motivation.}
Primality testing and factorization arise in cryptography agents, math assistants, and puzzle solvers. Recognizing small primes is easy, but testing primality of large numbers or factoring them requires systematic trial division or more advanced algorithms that exceed mental capacity.

\textbf{Tools.}
\begin{itemize}
    \item \texttt{is\_prime(n)}: Tests whether $n$ is prime and returns a boolean.
    \item \texttt{nth\_prime(n)}: Returns the $n$-th prime number.
    \item \texttt{factorize(n)}: Returns the complete prime factorization of $n$ as a string (e.g., ``$2 \times 3 \times 5$'').
\end{itemize}

\textbf{Answer format.} Boolean, factor string, or integer.

\textbf{Difficulty levels.}
\begin{itemize}
    \item \textbf{Easy}: Small primes and factorizations.\\
    \emph{Example}: ``Is 17 a prime number?'' $\to$ \textbf{True}
    \item \textbf{Medium}: 3-digit numbers.\\
    \emph{Example}: ``What is the 50th prime number?'' $\to$ \textbf{229}
    \item \textbf{Hard}: 5--6 digit numbers.\\
    \emph{Example}: ``What is the prime factorization of 8191?'' $\to$ \textbf{8191} (it is prime)
\end{itemize}

\subsection{Category B: Knowledge boundary}

These environments test whether the model can assess what information exists in its own parameters. The model must judge whether it possesses the factual knowledge needed to answer, a fundamentally different self-assessment from computational feasibility.

\subsubsection{RetrieverEnv}

\textbf{Motivation.}
Research agents and question-answering systems routinely need to look up facts from external corpora. The key self-assessment is whether the model already knows the answer from pretraining or needs to search. This environment is unique in requiring a two-step retrieval process: first searching for relevant documents, then reading the full content.

\textbf{Tools.}
\begin{itemize}
    \item \texttt{search\_corpus(query, top\_k)}: Searches a document corpus by keyword matching against document titles. Returns metadata and a short snippet (first 100 characters) for the top-$k$ results, but \emph{not} the full document text.
    \item \texttt{read\_doc(doc\_id)}: Retrieves the full text of a specific document by its ID, including title, content, and word count.
\end{itemize}
This is the only environment requiring two tool calls: the model must first call \texttt{search\_corpus} to identify the relevant document, then call \texttt{read\_doc} to obtain the full text containing the answer.

\textbf{Answer format.} Exact string (name, number, or phrase).

\textbf{Difficulty levels.}
\begin{itemize}
    \item \textbf{Easy}: Well-known facts across 10 categories (capitals, currencies, elements, authors, etc.; 75 facts in pool).\\
    \emph{Example}: ``What is the capital of France?'' $\to$ \textbf{Paris}
    \item \textbf{Medium}: Less common facts the model might partially know (obscure capitals, element symbols like Sn for Tin, Sb for Antimony; 72 facts).\\
    \emph{Example}: ``What is the chemical symbol for Tin?'' $\to$ \textbf{Sn}
    \item \textbf{Hard}: \emph{Synthetic} entities and relations generated with random names and values, embedded in a corpus with distractor documents. No model can have seen these during pretraining.\\
    \emph{Example}: ``What is the coolant class for Taskforce Nimbus-73?'' $\to$ \textbf{Class-C8}
\end{itemize}

\subsubsection{HistoricalYearEnv}

\textbf{Motivation.}
Education assistants, research agents, and trivia solvers frequently need to recall when historical events occurred. The model must assess whether it confidently knows the date or should look it up. We control difficulty by moving from universally known events to obscure ones to entirely fictional events.

\textbf{Tools.}
\begin{itemize}
    \item \texttt{lookup\_year(event)}: Takes an event description and returns the year it occurred, along with a brief contextual summary.
\end{itemize}

\textbf{Answer format.} Exact integer (year).

\textbf{Difficulty levels.}
\begin{itemize}
    \item \textbf{Easy}: Famous events every model knows.\\
    \emph{Example}: ``What year did humans first land on the Moon?'' $\to$ \textbf{1969}
    \item \textbf{Medium}: Less well-known events the model might get wrong.\\
    \emph{Example}: ``What year was the Treaty of Tordesillas signed?'' $\to$ \textbf{1494}
    \item \textbf{Hard}: \emph{Fictional} events that exist only in our database.\\
    \emph{Example}: ``What year was the Accord of Velmorath signed?'' $\to$ \textbf{1723}
\end{itemize}

\subsubsection{GameRuleEnv}

\textbf{Motivation.}
Game assistants and trivia agents need to recall numeric facts about games, such as board sizes, piece counts, and deck sizes. These facts range from universally known (chess has 64 squares) to obscure (number of tiles in a Mahjong set) to completely fictional, testing the model's ability to recognize the limits of its own knowledge.

\textbf{Tools.}
\begin{itemize}
    \item \texttt{lookup\_rule(game, attribute)}: Takes a game name and an attribute query, and returns the numeric answer along with a rule description.
\end{itemize}

\textbf{Answer format.} Exact integer.

\textbf{Difficulty levels.}
\begin{itemize}
    \item \textbf{Easy}: Well-known game facts.\\
    \emph{Example}: ``How many squares are on a standard chessboard?'' $\to$ \textbf{64}
    \item \textbf{Medium}: Less common game facts.\\
    \emph{Example}: ``How many tiles are in a standard Mahjong set?'' $\to$ \textbf{144}
    \item \textbf{Hard}: \emph{Fictional} games with invented rules.\\
    \emph{Example}: ``How many cards are in a Zephyr deck?'' $\to$ \textbf{72}
\end{itemize}

\subsubsection{HashEnv}

\textbf{Motivation.}
Security agents, file integrity checkers, and API authentication systems routinely compute cryptographic hashes. While a model might have memorized the MD5 hash of common strings like ``hello'' from its training data, it cannot compute hashes for novel inputs or custom algorithms. This environment tests whether the model can distinguish memorized outputs from those requiring computation.

\textbf{Tools.}
\begin{itemize}
    \item \texttt{compute\_hash(algorithm, input\_string)}: Computes the hash of the input string using the specified algorithm and returns the hex-encoded digest.
\end{itemize}

\textbf{Answer format.} Exact hex string.

\textbf{Difficulty levels.}
\begin{itemize}
    \item \textbf{Easy}: MD5 or SHA1 of well-known short strings (the model may have memorized these).\\
    \emph{Example}: ``What is the MD5 hash of `hello'?'' $\to$ \textbf{5d41402abc4b2a76b9719d911017c592}
    \item \textbf{Medium}: SHA256/SHA1/MD5 of short phrases less likely memorized.\\
    \emph{Example}: ``What is the SHA1 hash of `machine learning'?''
    \item \textbf{Hard}: 5 \emph{custom} hash algorithms (\texttt{fnv1a\_custom}, \texttt{djb2\_custom}, \texttt{sdbm\_custom}, \texttt{murmur\_custom}, \texttt{jenkins\_custom}) with non-standard primes and offsets. No model can know the outputs of these algorithms.\\
    \emph{Example}: ``What is the MURMUR\_CUSTOM hash of `xK9mQ2'?''
\end{itemize}

\subsubsection{DecodingEnv}

\textbf{Motivation.}
Communication agents and puzzle solvers encounter various encoding and cipher schemes. Morse code for ``SOS'' is universally known, and ROT13 is a simple well-known transformation, but custom substitution ciphers with arbitrary mappings cannot be decoded without access to the cipher definition. This environment tests the boundary between known and unknown encoding schemes.

\textbf{Tools.}
\begin{itemize}
    \item \texttt{decode(scheme, ciphertext)}: Decodes the ciphertext using the specified scheme and returns the plaintext.
    \item \texttt{encode(scheme, plaintext)}: Encodes the plaintext using the specified scheme and returns the ciphertext.
\end{itemize}

\textbf{Answer format.} Exact string.

\textbf{Difficulty levels.}
\begin{itemize}
    \item \textbf{Easy}: Morse code for short well-known words and ROT13.\\
    \emph{Example}: ``Encode `SOS' in Morse code.'' $\to$ \textbf{... -{-}- ...}
    \item \textbf{Medium}: Caesar cipher with various shifts and Morse for longer words.\\
    \emph{Example}: ``Decode `NWTPYE' using Caesar cipher with shift 11.'' $\to$ \textbf{CIPHER}
    \item \textbf{Hard}: 4 \emph{custom} substitution ciphers (\texttt{scramble1}, \texttt{scramble2}, \texttt{alpha7}, \texttt{reverse}) with arbitrary letter mappings that the model has never seen.\\
    \emph{Example}: ``Decode `KFPQA' using the scramble1 cipher.'' $\to$ \textbf{HELLO}
\end{itemize}

\subsection{Category C: Execution tracking}

These environments test whether the model can assess its own reliability when tracing sequential procedures. The model knows the algorithm and has all the information; the question is whether it can execute the steps without error.

\subsubsection{ListManipulationEnv}

\textbf{Motivation.}
Data processing agents, database operations, and ETL pipelines frequently perform list transformations, including inserting, removing, sorting, and reversing elements. Tracking a single operation on a short list is trivial, but applying operations to larger lists or 2D arrays (where row and column axes must be tracked simultaneously) quickly exceeds reliable mental execution.

\textbf{Tools.}
\begin{itemize}
    \item \texttt{append(list, value)}: Appends a value to the end of the list and returns the updated list.
    \item \texttt{remove(list, index)}: Removes the element at the given index and returns the updated list.
    \item \texttt{insert(list, index, value)}: Inserts a value at the given index and returns the updated list.
    \item \texttt{sort(list, axis)}: Sorts the list (or sorts a 2D list along the specified axis) and returns the result.
    \item \texttt{reverse(list)}: Reverses the list and returns the result.
\end{itemize}

\textbf{Answer format.} Exact list in Python format.

\textbf{Difficulty levels.}
\begin{itemize}
    \item \textbf{Easy}: 1D list of 3--5 small integers (1--40), single operation.\\
    \emph{Example}: ``Initial [7, 19, 29]. Apply insert(index=2, value=36). Return final list.'' $\to$ \textbf{[7, 19, 36, 29]}
    \item \textbf{Medium}: 1D list of 6--10 medium integers (40--260), single operation.\\
    \emph{Example}: ``Initial [86, 197, 199, 232, 66, 53, 234]. Apply sort(). Return final list.'' $\to$ \textbf{[53, 66, 86, 197, 199, 232, 234]}
    \item \textbf{Hard}: 2D list (matrix) of large integers (300--5000), operations along row or column axis.\\
    \emph{Example}: ``Initial [[2063, 3740, ...], [4252, 3661, ...]]. Apply sort(axis=0). Return final 2D list.''
\end{itemize}

\subsubsection{DateTimeEnv}

\textbf{Motivation.}
Scheduling agents, calendar assistants, and deadline trackers must perform date arithmetic, such as counting days between dates, adding durations, and determining days of the week. Simple within-month counting is easy, but calculations that cross month boundaries, handle leap years, or require day-of-week computation for arbitrary dates involve enough edge cases that mental execution becomes unreliable.

\textbf{Tools.}
\begin{itemize}
    \item \texttt{date\_add(date, days)}: Adds a number of days to a date and returns the resulting date in YYYY-MM-DD format.
    \item \texttt{date\_diff(date1, date2)}: Computes the number of days between two dates.
    \item \texttt{day\_of\_week(date)}: Returns the day of the week for a given date.
\end{itemize}

\textbf{Answer format.} Exact number, date string (YYYY-MM-DD), or day name.

\textbf{Difficulty levels.}
\begin{itemize}
    \item \textbf{Easy}: Simple counting within one month.\\
    \emph{Example}: ``How many days between January 3 and January 18?'' $\to$ \textbf{15}
    \item \textbf{Medium}: Crossing month/year boundaries, leap years.\\
    \emph{Example}: ``How many days between February 25 and March 10, 2024?'' $\to$ \textbf{14} (2024 is a leap year)
    \item \textbf{Hard}: Day-of-week for arbitrary dates, multi-year calculations.\\
    \emph{Example}: ``What day of the week is August 15, 2027?'' $\to$ \textbf{Sunday}
\end{itemize}

\subsubsection{CodeExecutorEnv}

\textbf{Motivation.}
Coding assistants and code review agents are frequently asked to predict the output of code snippets. Simple expressions are trivial, but code involving loops, recursion, or dynamic programming requires tracing many iterations where errors compound. This environment tests whether the model can recognize when code is too complex to trace mentally.

\textbf{Tools.}
\begin{itemize}
    \item \texttt{run\_python(code)}: Executes a Python code snippet in a sandboxed environment and returns the captured stdout output.
\end{itemize}

\textbf{Answer format.} Exact stdout string.

\textbf{Difficulty levels.}
\begin{itemize}
    \item \textbf{Easy}: Simple one-line expressions.\\
    \emph{Example}: ``What is the output of: \texttt{print(len('hello'))}'' $\to$ \textbf{5}
    \item \textbf{Medium}: Short loops, list comprehensions, string operations.\\
    \emph{Example}: ``What is the output of: \texttt{print(sum(x**2 for x in range(1,6)))}'' $\to$ \textbf{55}
    \item \textbf{Hard}: Recursion, dynamic programming, Collatz sequences: code with 10--30+ iterations where mental tracing reliably fails.\\
    \emph{Example}: ``What is the output of: \texttt{n=27; steps=0; while n!=1: n=n//2 if n\%2==0 else 3*n+1; steps+=1; print(steps)}'' $\to$ \textbf{111}
\end{itemize}

\subsubsection{ScheduleEnv}

\textbf{Motivation.}
Meeting schedulers and resource booking agents must find free time slots among existing appointments. With 2--3 meetings, checking for conflicts is easy. With 10+ overlapping meetings and specific duration constraints, mentally tracking all intervals to find available slots becomes error-prone.

\textbf{Tools.}
\begin{itemize}
    \item \texttt{find\_free\_slot(meetings, duration, start, end)}: Finds available time slots of the specified duration within the given time range, considering all existing meetings.
    \item \texttt{check\_conflict(meetings, new\_meeting)}: Checks whether a proposed meeting conflicts with any existing meetings and returns a boolean.
    \item \texttt{list\_meetings(meetings)}: Returns a formatted summary of all meetings sorted by start time.
\end{itemize}

\textbf{Answer format.} Exact time slot or boolean.

\textbf{Difficulty levels.}
\begin{itemize}
    \item \textbf{Easy}: 2--3 meetings, find a free slot.\\
    \emph{Example}: ``Meetings: 9:00--10:00, 14:00--15:00. Is there a free 1-hour slot between 10:00 and 14:00?'' $\to$ \textbf{Yes}
    \item \textbf{Medium}: 6--10 meetings, find all free slots.\\
    \emph{Example}: ``Given 8 meetings, list all free 30-min slots between 9:00 and 17:00.''
    \item \textbf{Hard}: 15+ meetings with constraints, requiring the model to trace all intervals.\\
    \emph{Example}: ``Given 15 meetings, find the first available 1-hour slot.''
\end{itemize}

\subsubsection{RegexMatchEnv}

\textbf{Motivation.}
Log parsing agents, data extraction pipelines, and text processing tools rely on regular expressions. While simple patterns like \texttt{\textbackslash d+} on short strings are easy to mentally evaluate, complex patterns with overlapping matches, lookahead assertions, or backreferences on long strings require the model to simulate a regex engine, a sequential process that quickly exceeds reliable mental execution.

\textbf{Tools.}
\begin{itemize}
    \item \texttt{regex\_match(pattern, text, operation)}: Applies the specified regex operation (\texttt{findall}, \texttt{match}, \texttt{search}, \texttt{sub}) to the text and returns the result.
\end{itemize}

\textbf{Answer format.} Exact Python list or matched string.

\textbf{Difficulty levels.}
\begin{itemize}
    \item \textbf{Easy}: Simple character classes and quantifiers on short text.\\
    \emph{Example}: ``What does \texttt{re.findall(r'\textbackslash d+', `abc123def456')} return?'' $\to$ \textbf{['123', '456']}
    \item \textbf{Medium}: Groups, lookahead/lookbehind, alternation on moderate text.\\
    \emph{Example}: ``What does \texttt{re.findall(r'(\textbackslash w+)@(\textbackslash w+)\textbackslash.(\textbackslash w+)', `user@example.com admin@test.org')} return?'' $\to$ \textbf{[('user', 'example', 'com'), ('admin', 'test', 'org')]}
    \item \textbf{Hard}: Complex patterns with overlapping matches on 60+ character strings, requiring the model to simulate regex engine backtracking.\\
    \emph{Example}: ``What does \texttt{re.findall(r'(?=([a-z]\{3\}))', `abcdbcdabcdebcde...')} return?''
\end{itemize}

\subsection{Multi-hop environments}

In addition to the 15 single-hop environments, \benchmark{} includes 3 multi-hop environments that require chains of 3 dependent tool calls following the pattern $x \to y \to z$: each hop's output is needed as input for the next. These test whether models can assess tool necessity across a sequence of dependent operations.

All three multi-hop environments follow the same structure: the model executes hop~1 to obtain~$x$, uses~$x$ as input to hop~2 to obtain~$y$, and uses~$y$ as input to hop~3 to obtain the final answer~$z$. Each hop reuses the same tools as its single-hop counterpart. Difficulty scales identically (easy = solvable mentally, hard = requires tools), but now the model must make a tool-call decision at each hop.

\subsubsection{ChainedCalculatorEnv (Category A)}

Three chained arithmetic computations where each expression depends on the previous result. Uses \texttt{evaluate\_expression}.

\emph{Easy}: ``First compute $x = 40 - 10$. Then compute $y = x + 5$. Finally compute $z = y - 19$. Return $z$.'' $\to$ $x{=}30, y{=}35, z{=}16$.\\
\emph{Hard}: ``First compute $x = 808522010435 - 8197325888$. Then compute $y = x + 17046220916$. Finally compute $z = y \bmod 2343374$. Return $z$.'' $\to$ $x{=}800324684547, y{=}817370905463, z{=}2054263$.

\subsubsection{ChainedRetrieverEnv (Category B)}

Three chained knowledge lookups where the answer to each question determines what to ask next. The corpus contains target documents for all three hops plus distractors (~10 docs). Uses \texttt{search\_corpus} and \texttt{read\_doc}.

\emph{Easy}: ``What is the longest river in the world? ($x{=}$Nile). Through how many countries does $x$ flow? ($y{=}$11). Into which sea does the river that passes through $y$ countries empty? ($z{=}$Mediterranean Sea).''\\
\emph{Hard}: ``What river formed the Central Lowlands? ($x{=}$the Ironflow River). What gives $x$ its name? ($y{=}$iron-rich sediments). What color is the water of the river that carries $y$? ($z{=}$reddish-brown).'' All entities are fictional and exist only in the provided corpus.

\subsubsection{ChainedCodeExecutorEnv (Category C)}

Three chained code executions where each script's output is the input to the next. Uses \texttt{run\_code}.

\emph{Easy}: ``Run \texttt{print(17+6)} ($x{=}23$). Run \texttt{print(x*3)} ($y{=}69$). Run \texttt{print(y-7)} ($z{=}62$).''\\
\emph{Hard}: ``Run coin-change DP for amount=27 with coins [1,5,10] ($x{=}5$). Sum even Fibonacci numbers up to $2x$ ($y{=}44$). Compute LIS of an array with first element replaced by $y\bmod50$ ($z{=}7$).'' Each hop involves loops, recursion, or DP.

\subsection{Difficulty validation}

To validate that our difficulty levels create a meaningful decision boundary, we evaluate all six models in a no-tool setting where tools are unavailable and the model must answer directly. Table~\ref{tab:difficulty_validation} reports the results across all 18 environments (single-hop and multi-hop combined). Averaged across models, easy tasks are solvable 69.4\% of the time, medium tasks 54.4\%, and hard tasks only 15.5\%, confirming that the difficulty levels behave as intended.

\begin{table}[H]
\caption{No-tool accuracy (\%) by difficulty level across all 18 environments (1{,}080 train / 2{,}700 test). Each model is evaluated without tool access to validate that easy tasks are largely solvable directly, medium tasks are partially solvable, and hard tasks require tools.}
\label{tab:difficulty_validation}
\centering
\small
\begin{tabular}{l cccc cccc}
\toprule
& \multicolumn{4}{c}{Train (1{,}080 tasks)} & \multicolumn{4}{c}{Test (2{,}700 tasks)} \\
\cmidrule(lr){2-5} \cmidrule(lr){6-9}
Model & Easy & Med & Hard & Avg & Easy & Med & Hard & Avg \\
\midrule
Qwen3-1.7B     & 34.7 & 26.1 &  4.7 & 21.9 & 37.0 & 25.2 &  6.3 & 22.9 \\
Qwen3-4B-Inst. & 73.6 & 64.4 & 21.9 & 53.3 & 74.4 & 61.2 & 20.4 & 52.0 \\
Qwen3-14B      & 81.1 & 61.1 & 17.5 & 53.2 & 77.9 & 58.1 & 15.2 & 50.4 \\
Qwen3-32B      & 83.3 & 69.4 & 20.3 & 57.7 & 82.8 & 69.1 & 21.3 & 57.7 \\
Llama-3.1-8B   & 69.7 & 48.6 &  9.2 & 42.5 & 69.8 & 48.9 &  9.4 & 42.7 \\
Llama-3.3-70B  & 75.8 & 65.0 & 22.2 & 54.4 & 74.7 & 63.7 & 20.6 & 53.0 \\
\midrule
Average         & 69.7 & 55.8 & 16.0 & 47.2 & 69.4 & 54.4 & 15.5 & 46.4 \\
\bottomrule
\end{tabular}
\end{table}

\newpage
\section{Detailed single-hop results}
\label{app:full_reduction}

This section provides the complete per-model results for the 15 single-hop environments that are summarized in the main text figures and tables.

\paragraph{Accuracy cost per saved call.}
Table~\ref{tab:full_reduction} breaks down the accuracy cost per saved call ($\frac{\Delta\text{Acc}}{-\Delta\text{TC}}$) from Table~\ref{tab:adaptive_reduction} by model. For each model, we compare Sparse~(S), Sparse + Reason-then-Act, and \method{} ($\tau$=0.5), all relative to Default~($\star$, Prompt-only).

\begin{table}[H]
\caption{Per-model accuracy cost per saved call ($\tau$=0.5). All deltas relative to Default~($\star$, Prompt-only). More negative = each saved call is more costly. \textbf{Bold} = best per model in each column.}
\label{tab:full_reduction}
\centering
\resizebox{\textwidth}{!}{%
\scriptsize
\setlength{\tabcolsep}{2.5pt}
\begin{tabular}{ll rrr rrr rrr rrr}
\toprule
& & \multicolumn{3}{c}{Easy} & \multicolumn{3}{c}{Medium} & \multicolumn{3}{c}{Hard} & \multicolumn{3}{c}{Avg} \\
\cmidrule(lr){3-5} \cmidrule(lr){6-8} \cmidrule(lr){9-11} \cmidrule(lr){12-14}
Model & Method & $\Delta$Acc & $\Delta$TC & $\frac{\Delta\text{Acc}}{-\Delta\text{TC}}$ & $\Delta$Acc & $\Delta$TC & $\frac{\Delta\text{Acc}}{-\Delta\text{TC}}$ & $\Delta$Acc & $\Delta$TC & $\frac{\Delta\text{Acc}}{-\Delta\text{TC}}$ & $\Delta$Acc & $\Delta$TC & $\frac{\Delta\text{Acc}}{-\Delta\text{TC}}$ \\
\midrule
\multirow{3}{*}{Qwen3-1.7B}
  & Sparse (S)       & $-$17.7 & $-$0.59 & $-$29.9 & $-$18.4 & $-$0.39 & $-$46.7 & $-$18.0 & $-$0.48 & $-$37.7 & $-$18.0 & $-$0.49 & $-$37.1 \\
  & Sparse + Reason-then-Act   & $-$8.7  & $-$0.83 & $-$10.5 & $-$17.9 & $-$0.62 & $-$28.7 & $-$18.4 & $-$0.61 & $-$30.0 & $-$15.0 & $-$0.69 & $-$21.8 \\
  & \method{}        & \textbf{$-$0.9}  & $-$0.37 & \textbf{$-$2.4}  & \textbf{$+$0.2} & $-$0.23 & \textbf{$+$1.0} & \textbf{$+$0.9} & $-$0.17 & \textbf{$+$5.3} & \textbf{$+$0.1} & $-$0.26 & \textbf{$+$0.3} \\
\midrule
\multirow{3}{*}{Qwen3-4B-Inst.}
  & Sparse (S)       & $-$14.5 & $-$0.84 & $-$17.3 & $-$20.7 & $-$0.86 & $-$24.1 & $-$20.3 & $-$0.48 & $-$42.4 & $-$18.5 & $-$0.73 & $-$25.5 \\
  & Sparse + Reason-then-Act   & $-$14.5 & $-$0.86 & $-$16.9 & $-$22.4 & $-$0.90 & $-$24.8 & $-$13.0 & $-$0.35 & $-$36.6 & $-$16.6 & $-$0.70 & $-$23.6 \\
  & \method{}        & \textbf{$-$2.5}  & $-$0.49 & \textbf{$-$5.1}  & \textbf{$-$5.5}  & $-$0.51 & \textbf{$-$10.7} & \textbf{$+$6.0} & $-$0.08 & \textbf{$+$73.9} & \textbf{$-$0.7} & $-$0.36 & \textbf{$-$1.9} \\
\midrule
\multirow{3}{*}{Qwen3-14B}
  & Sparse (S)       & $-$8.8  & $-$0.59 & $-$14.9 & $-$12.9 & $-$0.53 & $-$24.3 & $-$27.3 & $-$0.47 & $-$58.4 & $-$16.3 & $-$0.53 & $-$30.8 \\
  & Sparse + Reason-then-Act   & $-$4.4  & $-$0.67 & $-$6.6  & $-$10.4 & $-$0.62 & $-$16.8 & $-$9.7  & $-$0.28 & $-$34.7 & $-$8.2 & $-$0.52 & $-$15.6 \\
  & \method{}        & \textbf{$-$0.4}  & $-$0.63 & \textbf{$-$0.6}  & \textbf{$-$3.6}  & $-$0.55 & \textbf{$-$6.5}  & \textbf{$+$0.2} & $-$0.13 & \textbf{$+$1.4} & \textbf{$-$1.3} & $-$0.44 & \textbf{$-$2.9} \\
\midrule
\multirow{3}{*}{Qwen3-32B}
  & Sparse (S)       & \textbf{$-$0.8}  & $-$0.38 & \textbf{$-$2.2}  & \textbf{$-$0.3}  & $-$0.20 & \textbf{$-$1.6}  & \textbf{$-$1.3}  & $-$0.08 & \textbf{$-$16.9} & \textbf{$-$0.8} & $-$0.22 & \textbf{$-$3.6} \\
  & Sparse + Reason-then-Act   & $-$4.6  & $-$0.80 & $-$5.8  & $-$8.8  & $-$0.56 & $-$15.8 & $-$5.3  & $-$0.21 & $-$25.1 & $-$6.2 & $-$0.52 & $-$11.9 \\
  & \method{}        & $-$2.1  & $-$0.83 & $-$2.5  & $-$6.4  & $-$0.66 & $-$9.7  & $-$3.6  & $-$0.18 & $-$20.1 & $-$4.0 & $-$0.56 & $-$7.2 \\
\midrule
\multirow{3}{*}{Llama-3.1-8B}
  & Sparse (S)       & \textbf{$+$2.7}  & $-$0.42 & \textbf{$+$6.4}  & \textbf{$+$2.8}  & $-$0.43 & \textbf{$+$6.4}  & \textbf{$+$0.4}  & $-$0.27 & \textbf{$+$1.6} & \textbf{$+$2.0} & $-$0.37 & \textbf{$+$5.3} \\
  & Sparse + Reason-then-Act   & $-$22.6 & $-$1.66 & $-$13.6 & $-$41.0 & $-$1.69 & $-$24.3 & $-$65.8 & $-$1.60 & $-$41.2 & $-$43.1 & $-$1.65 & $-$26.1 \\
  & \method{}        & $-$5.3  & $-$0.84 & $-$6.4  & $-$10.9 & $-$0.68 & $-$16.1 & $-$13.1 & $-$0.25 & $-$51.3 & $-$9.8 & $-$0.59 & $-$16.6 \\
\midrule
\multirow{3}{*}{Llama-3.3-70B}
  & Sparse (S)       & $+$1.6  & $-$0.51 & $+$3.2  & $+$2.0  & $-$0.41 & $+$4.8  & $-$0.2  & $-$0.34 & $-$0.5 & $+$1.1 & $-$0.42 & $+$2.7 \\
  & Sparse + Reason-then-Act   & $-$4.8  & $-$1.98 & $-$2.4  & $-$18.9 & $-$1.87 & $-$10.1 & $-$63.3 & $-$1.99 & $-$31.7 & $-$29.0 & $-$1.95 & $-$14.9 \\
  & \method{}        & \textbf{$+$4.8} & $-$0.78 & \textbf{$+$6.2} & \textbf{$+$6.0} & $-$0.61 & \textbf{$+$9.8} & \textbf{$+$4.8} & $-$0.62 & \textbf{$+$7.6} & \textbf{$+$5.2} & $-$0.67 & \textbf{$+$7.8} \\
\bottomrule
\end{tabular}%
}
\end{table}

\paragraph{Full single-hop results.}
Table~\ref{tab:singlehop_full} reports accuracy (\%) and total tool calls (TC) for all 6 models across all prompt modes, reasoning settings, and probe thresholds on the 2{,}250-task single-hop test set.

\begin{table}[H]
\caption{Full single-hop results (2{,}250 test tasks, mean$\pm$std over 3 runs). F/D/N/S/X = Force/Default/Necessary/Sparse/No-tool. $\tau$ = probe threshold.}
\label{tab:singlehop_full}
\centering
\resizebox{\textwidth}{!}{%
\tiny
\setlength{\tabcolsep}{1.5pt}
\renewcommand{\arraystretch}{1.15}
\begin{tabular}{ll cc cc cc cc cc cc}
\toprule
& & \multicolumn{2}{c}{Qwen3-1.7B} & \multicolumn{2}{c}{Qwen3-4B} & \multicolumn{2}{c}{Qwen3-14B} & \multicolumn{2}{c}{Qwen3-32B} & \multicolumn{2}{c}{Llama-8B} & \multicolumn{2}{c}{Llama-70B} \\
\cmidrule(lr){3-4} \cmidrule(lr){5-6} \cmidrule(lr){7-8} \cmidrule(lr){9-10} \cmidrule(lr){11-12} \cmidrule(lr){13-14}
& & Acc & TC & Acc & TC & Acc & TC & Acc & TC & Acc & TC & Acc & TC \\
\midrule
\multirow{5}{*}{\rotatebox{90}{Prompt-only}}
& F & 86.8{\tiny$\pm$.2} & 3120{\tiny$\pm$32} & 92.0{\tiny$\pm$.0} & 2435{\tiny$\pm$5} & 93.7{\tiny$\pm$.1} & 2421{\tiny$\pm$7} & 92.8{\tiny$\pm$.1} & 2559{\tiny$\pm$9} & 78.7{\tiny$\pm$.4} & 4081{\tiny$\pm$40} & 78.3{\tiny$\pm$.3} & 5302{\tiny$\pm$55} \\
& D & 88.2{\tiny$\pm$.1} & 2709{\tiny$\pm$22} & 89.2{\tiny$\pm$.2} & 2118{\tiny$\pm$9} & 93.7{\tiny$\pm$.0} & 2211{\tiny$\pm$2} & 94.1{\tiny$\pm$.3} & 2404{\tiny$\pm$20} & 79.5{\tiny$\pm$.6} & 3708{\tiny$\pm$55} & 83.1{\tiny$\pm$.2} & 4377{\tiny$\pm$7} \\
& N & 87.2{\tiny$\pm$.3} & 2787{\tiny$\pm$13} & 85.7{\tiny$\pm$.1} & 1852{\tiny$\pm$3} & 93.5{\tiny$\pm$.2} & 2175{\tiny$\pm$5} & 93.5{\tiny$\pm$.2} & 2394{\tiny$\pm$23} & 77.7{\tiny$\pm$.6} & 3548{\tiny$\pm$51} & 84.4{\tiny$\pm$.1} & 3988{\tiny$\pm$32} \\
& S & 70.2{\tiny$\pm$.2} & 1611{\tiny$\pm$20} & 70.7{\tiny$\pm$.3} & 484{\tiny$\pm$10} & 77.3{\tiny$\pm$.2} & 1020{\tiny$\pm$3} & 93.3{\tiny$\pm$.3} & 1912{\tiny$\pm$10} & 81.4{\tiny$\pm$.4} & 2868{\tiny$\pm$3} & 84.2{\tiny$\pm$.1} & 3431{\tiny$\pm$15} \\
& X & 26.3{\tiny$\pm$.1} & 293{\tiny$\pm$13} & 50.8{\tiny$\pm$.5} & 121{\tiny$\pm$14} & 51.2{\tiny$\pm$.2} & 85{\tiny$\pm$5} & 66.8{\tiny$\pm$.5} & 451{\tiny$\pm$15} & 77.6{\tiny$\pm$.2} & 3391{\tiny$\pm$53} & 82.5{\tiny$\pm$.3} & 3615{\tiny$\pm$39} \\
\midrule
\multirow{5}{*}{\rotatebox{90}{Reason-then-Act}}
& F & 86.7{\tiny$\pm$.5} & 2475{\tiny$\pm$19} & 90.9{\tiny$\pm$.1} & 1651{\tiny$\pm$6} & 93.4{\tiny$\pm$.1} & 2172{\tiny$\pm$9} & 93.7{\tiny$\pm$.1} & 2150{\tiny$\pm$12} & 29.3{\tiny$\pm$.2} & 9{\tiny$\pm$8} & 42.4{\tiny$\pm$.2} & 0{\tiny$\pm$0} \\
& D & 84.7{\tiny$\pm$.4} & 1923{\tiny$\pm$27} & 83.4{\tiny$\pm$.2} & 1024{\tiny$\pm$0} & 92.0{\tiny$\pm$.2} & 1589{\tiny$\pm$3} & 92.9{\tiny$\pm$.4} & 1823{\tiny$\pm$6} & 31.2{\tiny$\pm$.9} & 2{\tiny$\pm$0} & 47.9{\tiny$\pm$.5} & 0{\tiny$\pm$0} \\
& N & 84.0{\tiny$\pm$.7} & 1871{\tiny$\pm$30} & 83.4{\tiny$\pm$.4} & 1027{\tiny$\pm$7} & 92.4{\tiny$\pm$.2} & 1634{\tiny$\pm$13} & 93.2{\tiny$\pm$.1} & 1845{\tiny$\pm$7} & 32.0{\tiny$\pm$.1} & 11{\tiny$\pm$3} & 47.6{\tiny$\pm$.5} & 0{\tiny$\pm$0} \\
& S & 73.2{\tiny$\pm$.6} & 1161{\tiny$\pm$22} & 72.5{\tiny$\pm$.7} & 535{\tiny$\pm$15} & 85.5{\tiny$\pm$.3} & 1034{\tiny$\pm$12} & 87.9{\tiny$\pm$.2} & 1231{\tiny$\pm$20} & 36.3{\tiny$\pm$.4} & 0{\tiny$\pm$0} & 54.1{\tiny$\pm$.2} & 0{\tiny$\pm$0} \\
& X & 71.1{\tiny$\pm$.3} & 1117{\tiny$\pm$19} & 64.5{\tiny$\pm$.4} & 315{\tiny$\pm$12} & 83.5{\tiny$\pm$.4} & 928{\tiny$\pm$9} & 74.3{\tiny$\pm$.3} & 596{\tiny$\pm$19} & 34.5{\tiny$\pm$.3} & 6{\tiny$\pm$2} & 54.8{\tiny$\pm$.4} & 0{\tiny$\pm$0} \\
\midrule
\multirow{5}{*}{\rotatebox{90}{\method{}}}
& $\tau$=.1 & 88.8{\tiny$\pm$.2} & 2512{\tiny$\pm$16} & 91.1{\tiny$\pm$.3} & 2216{\tiny$\pm$10} & 94.3{\tiny$\pm$.1} & 2128{\tiny$\pm$9} & 94.0{\tiny$\pm$.1} & 1996{\tiny$\pm$9} & 69.2{\tiny$\pm$.6} & 3027{\tiny$\pm$20} & 88.4{\tiny$\pm$.4} & 2976{\tiny$\pm$58} \\
& $\tau$=.3 & 89.0{\tiny$\pm$.3} & 2507{\tiny$\pm$36} & 90.4{\tiny$\pm$.4} & 1707{\tiny$\pm$5} & 94.1{\tiny$\pm$.2} & 1509{\tiny$\pm$4} & 93.2{\tiny$\pm$.1} & 1493{\tiny$\pm$10} & 68.9{\tiny$\pm$.2} & 2770{\tiny$\pm$33} & 88.6{\tiny$\pm$.2} & 2902{\tiny$\pm$31} \\
& $\tau$=.5 & 88.3{\tiny$\pm$.2} & 2128{\tiny$\pm$18} & 88.5{\tiny$\pm$.2} & 1309{\tiny$\pm$1} & 92.4{\tiny$\pm$.2} & 1227{\tiny$\pm$5} & 90.1{\tiny$\pm$.5} & 1155{\tiny$\pm$22} & 69.7{\tiny$\pm$.4} & 2381{\tiny$\pm$28} & 88.3{\tiny$\pm$.1} & 2871{\tiny$\pm$49} \\
& $\tau$=.7 & 81.6{\tiny$\pm$.3} & 1415{\tiny$\pm$19} & 84.8{\tiny$\pm$.2} & 1026{\tiny$\pm$9} & 85.8{\tiny$\pm$.1} & 907{\tiny$\pm$6} & 82.3{\tiny$\pm$.3} & 896{\tiny$\pm$16} & 66.5{\tiny$\pm$.2} & 2146{\tiny$\pm$30} & 88.6{\tiny$\pm$.5} & 2828{\tiny$\pm$9} \\
& $\tau$=.9 & 47.9{\tiny$\pm$.5} & 293{\tiny$\pm$15} & 74.7{\tiny$\pm$.3} & 657{\tiny$\pm$4} & 66.0{\tiny$\pm$.6} & 347{\tiny$\pm$8} & 71.5{\tiny$\pm$.4} & 604{\tiny$\pm$14} & 61.7{\tiny$\pm$.4} & 1753{\tiny$\pm$23} & 89.2{\tiny$\pm$.2} & 2804{\tiny$\pm$10} \\
\bottomrule
\end{tabular}%
}
\end{table}

\newpage
\section{Multi-hop evaluation}
\label{app:multihop}

We evaluate \method{} on the 3 multi-hop environments (ChainedCalculatorEnv, ChainedRetrieverEnv, ChainedCodeExecutorEnv), each requiring a chain of 3 dependent tool calls. The probe is trained on the 180-task multi-hop training set and evaluated on the 450-task test set.

\paragraph{Summary.}
Table~\ref{tab:multihop_summary} compares the best baseline (highest accuracy among all Prompt-only and Reason-then-Act settings that achieve at least 20\% tool-call reduction) against the best probe threshold. On Qwen3-4B, the probe achieves higher accuracy (85.3\% vs.\ 83.9\%) with 75\% fewer tool calls compared to the best baseline's 63\%. On Qwen3-32B, the probe reduces calls by 55\% while the best baseline only achieves 20\%. On Llama models, the probe increases tool calls but also substantially increases accuracy (Llama-3.1-8B: 40.2$\to$60.2\%, Llama-3.3-70B: 62.4$\to$80.3\%), indicating that the probe correctly identifies these multi-hop tasks as genuinely requiring tools and steers the model toward necessary calls that the Default setting was missing.

\begin{table}[H]
\caption{Multi-hop summary (450 test tasks). $\Delta$TC relative to Default. \textbf{Bold} = larger reduction.}
\label{tab:multihop_summary}
\centering
\small
\setlength{\tabcolsep}{3pt}
\begin{tabular}{l c cc ccc ccc}
\toprule
& & \multicolumn{2}{c}{Default} & \multicolumn{3}{c}{Best Baseline} & \multicolumn{3}{c}{\method{}} \\
\cmidrule(lr){3-4} \cmidrule(lr){5-7} \cmidrule(lr){8-10}
Model & $N$ & Acc & TC & Acc & TC & $\Delta$TC & Acc & TC & $\Delta$TC \\
\midrule
Qwen3-1.7B   & 450 & 21.2 & 1180 & \textbf{60.6} & 175 & $-$85\% & 59.2 & 85 & \textbf{$-$93\%} \\
Qwen3-4B     & 450 & 82.1 & 1719 & 83.9 & 636 & $-$63\% & \textbf{85.3} & 437 & \textbf{$-$75\%} \\
Qwen3-14B    & 450 & 87.5 & 1503 & 85.7 & 789 & \textbf{$-$47\%} & \textbf{86.2} & 996 & $-$34\% \\
Qwen3-32B    & 450 & 88.9 & 1634 & 88.6 & 1306 & $-$20\% & \textbf{89.0} & 727 & \textbf{$-$55\%} \\
Llama-3.1-8B & 450 & 40.2 & 1005 & 41.3 & 595 & \textbf{$-$41\%} & \textbf{60.2} & 1361 & $+$35\% \\
Llama-3.3-70B& 450 & 62.4 & 985 & 67.6 & 347 & \textbf{$-$65\%} & \textbf{80.3} & 1789 & $+$82\% \\
\bottomrule
\end{tabular}
\end{table}

\paragraph{Probe quality.}
Table~\ref{tab:multihop_auroc} reports probe AUROC on the multi-hop test set. The probe achieves AUROC 0.84--0.97 across models, with Qwen3-4B reaching 0.966, confirming that tool necessity remains linearly decodable even for chained tasks. Llama-3.3-70B has the lowest AUROC (0.804), likely because this model's Default setting already under-calls tools on multi-hop tasks (TC=985 for 450 three-hop tasks), making the binary label noisier. Despite the lower AUROC, the probe still substantially improves accuracy on this model (62.4\% $\to$ 80.3\%).

\begin{table}[H]
\caption{Multi-hop probe AUROC and classification accuracy.}
\label{tab:multihop_auroc}
\centering
\small
\begin{tabular}{lcc}
\toprule
Model & AUROC & Accuracy \\
\midrule
Qwen3-1.7B   & 0.839 & 0.796 \\
Qwen3-4B     & 0.966 & 0.947 \\
Qwen3-14B    & 0.906 & 0.822 \\
Qwen3-32B    & 0.944 & 0.873 \\
Llama-3.1-8B & 0.895 & 0.829 \\
Llama-3.3-70B& 0.804 & 0.729 \\
\bottomrule
\end{tabular}
\end{table}

\paragraph{Full results.}
Table~\ref{tab:multihop_full} reports all settings. Several patterns emerge beyond the summary table. First, Reason-then-Act is particularly effective on Qwen3-1.7B multi-hop tasks (Sparse+R-t-A reaches 60.6\% vs.\ Sparse Prompt-only's 41.3\%), suggesting that explicit reasoning helps small models plan multi-step tool chains. Second, the Llama models exhibit the same reasoning collapse as on single-hop tasks: Reason-then-Act reduces tool calls to near zero on both Llama-3.1-8B (TC$\leq$5) and Llama-3.3-70B (TC=0), with accuracy dropping substantially. Third, \method{} shows a clear threshold--accuracy tradeoff: on Qwen3-4B, sweeping $\tau$ from 0.1 to 0.9 reduces TC from 1287 to 175 while accuracy decreases gradually from 82.1\% to 79.6\%, providing smooth control that baselines cannot achieve.

\begin{table}[H]
\caption{Full multi-hop results (450 test tasks, mean$\pm$std over 3 runs). F/D/N/S/X = Force/Default/Necessary/Sparse/No-tool. $\tau$ = probe threshold.}
\label{tab:multihop_full}
\centering
\resizebox{\textwidth}{!}{%
\tiny
\setlength{\tabcolsep}{1.5pt}
\renewcommand{\arraystretch}{1.15}
\begin{tabular}{ll cc cc cc cc cc cc}
\toprule
& & \multicolumn{2}{c}{Qwen3-1.7B} & \multicolumn{2}{c}{Qwen3-4B} & \multicolumn{2}{c}{Qwen3-14B} & \multicolumn{2}{c}{Qwen3-32B} & \multicolumn{2}{c}{Llama-8B} & \multicolumn{2}{c}{Llama-70B} \\
\cmidrule(lr){3-4} \cmidrule(lr){5-6} \cmidrule(lr){7-8} \cmidrule(lr){9-10} \cmidrule(lr){11-12} \cmidrule(lr){13-14}
& & Acc & TC & Acc & TC & Acc & TC & Acc & TC & Acc & TC & Acc & TC \\
\midrule
\multirow{5}{*}{\rotatebox{90}{Prompt-only}}
& F & 28.8{\tiny$\pm$.4} & 1438{\tiny$\pm$23} & 76.1{\tiny$\pm$.7} & 2002{\tiny$\pm$10} & 88.0{\tiny$\pm$.4} & 1787{\tiny$\pm$2} & 89.4{\tiny$\pm$.5} & 1797{\tiny$\pm$20} & 42.4{\tiny$\pm$1.1} & 1288{\tiny$\pm$20} & 77.8{\tiny$\pm$1.6} & 1243{\tiny$\pm$12} \\
& D & 21.2{\tiny$\pm$.6} & 1180{\tiny$\pm$37} & 82.1{\tiny$\pm$.2} & 1719{\tiny$\pm$14} & 87.5{\tiny$\pm$.8} & 1503{\tiny$\pm$4} & 88.9{\tiny$\pm$.8} & 1634{\tiny$\pm$22} & 40.2{\tiny$\pm$1.7} & 1005{\tiny$\pm$23} & 62.4{\tiny$\pm$.5} & 985{\tiny$\pm$32} \\
& N & 23.3{\tiny$\pm$1.0} & 1230{\tiny$\pm$21} & 84.3{\tiny$\pm$.5} & 1390{\tiny$\pm$17} & 87.0{\tiny$\pm$.5} & 1495{\tiny$\pm$10} & 85.4{\tiny$\pm$3.1} & 1586{\tiny$\pm$14} & 40.1{\tiny$\pm$.5} & 1017{\tiny$\pm$20} & 60.9{\tiny$\pm$1.3} & 793{\tiny$\pm$8} \\
& S & 41.3{\tiny$\pm$1.2} & 397{\tiny$\pm$16} & 83.6{\tiny$\pm$.6} & 83{\tiny$\pm$4} & 74.4{\tiny$\pm$1.2} & 496{\tiny$\pm$9} & 87.6{\tiny$\pm$.8} & 832{\tiny$\pm$28} & 41.3{\tiny$\pm$1.0} & 595{\tiny$\pm$42} & 67.6{\tiny$\pm$.6} & 347{\tiny$\pm$40} \\
& X & 26.7{\tiny$\pm$.9} & 108{\tiny$\pm$4} & 75.9{\tiny$\pm$.5} & 13{\tiny$\pm$2} & 67.4{\tiny$\pm$.9} & 384{\tiny$\pm$8} & 70.1{\tiny$\pm$.5} & 58{\tiny$\pm$10} & 34.3{\tiny$\pm$1.2} & 639{\tiny$\pm$36} & 36.7{\tiny$\pm$2.8} & 385{\tiny$\pm$22} \\
\midrule
\multirow{5}{*}{\rotatebox{90}{Reason-then-Act}}
& F & 51.9{\tiny$\pm$.5} & 841{\tiny$\pm$24} & 83.5{\tiny$\pm$.7} & 940{\tiny$\pm$4} & 87.3{\tiny$\pm$1.1} & 1326{\tiny$\pm$9} & 88.6{\tiny$\pm$.7} & 1306{\tiny$\pm$21} & 23.2{\tiny$\pm$.9} & 1{\tiny$\pm$0} & 56.0{\tiny$\pm$1.2} & 0{\tiny$\pm$0} \\
& D & 55.7{\tiny$\pm$2.4} & 345{\tiny$\pm$14} & 83.6{\tiny$\pm$.6} & 652{\tiny$\pm$7} & 85.7{\tiny$\pm$.4} & 789{\tiny$\pm$24} & 86.7{\tiny$\pm$1.1} & 1010{\tiny$\pm$21} & 29.3{\tiny$\pm$.9} & 2{\tiny$\pm$1} & 57.0{\tiny$\pm$1.7} & 0{\tiny$\pm$0} \\
& N & 56.4{\tiny$\pm$2.0} & 348{\tiny$\pm$23} & 83.9{\tiny$\pm$.5} & 636{\tiny$\pm$6} & 84.4{\tiny$\pm$.8} & 863{\tiny$\pm$26} & 87.1{\tiny$\pm$.5} & 1036{\tiny$\pm$21} & 29.6{\tiny$\pm$.8} & 5{\tiny$\pm$2} & 58.3{\tiny$\pm$.6} & 0{\tiny$\pm$0} \\
& S & 60.6{\tiny$\pm$.6} & 175{\tiny$\pm$13} & 81.9{\tiny$\pm$.9} & 253{\tiny$\pm$6} & 76.1{\tiny$\pm$.2} & 404{\tiny$\pm$16} & 82.0{\tiny$\pm$.8} & 565{\tiny$\pm$9} & 37.0{\tiny$\pm$2.7} & 4{\tiny$\pm$4} & 60.6{\tiny$\pm$1.2} & 0{\tiny$\pm$0} \\
& X & 59.9{\tiny$\pm$.9} & 206{\tiny$\pm$14} & 78.5{\tiny$\pm$.4} & 106{\tiny$\pm$3} & 73.5{\tiny$\pm$.6} & 301{\tiny$\pm$19} & 74.0{\tiny$\pm$1.1} & 294{\tiny$\pm$14} & 36.4{\tiny$\pm$.5} & 4{\tiny$\pm$4} & 60.3{\tiny$\pm$.7} & 0{\tiny$\pm$0} \\
\midrule
\multirow{5}{*}{\rotatebox{90}{\method{}}}
& $\tau$=.1 & 22.9{\tiny$\pm$.9} & 1197{\tiny$\pm$11} & 82.1{\tiny$\pm$.4} & 1287{\tiny$\pm$11} & 87.3{\tiny$\pm$.5} & 1483{\tiny$\pm$18} & 88.7{\tiny$\pm$.5} & 1481{\tiny$\pm$8} & 60.1{\tiny$\pm$1.5} & 1361{\tiny$\pm$9} & 80.1{\tiny$\pm$1.0} & 1856{\tiny$\pm$29} \\
& $\tau$=.3 & 24.4{\tiny$\pm$.5} & 1190{\tiny$\pm$31} & 85.3{\tiny$\pm$.6} & 437{\tiny$\pm$3} & 86.1{\tiny$\pm$.5} & 996{\tiny$\pm$13} & 89.0{\tiny$\pm$.6} & 727{\tiny$\pm$6} & 59.0{\tiny$\pm$.9} & 1291{\tiny$\pm$24} & 80.3{\tiny$\pm$1.2} & 1789{\tiny$\pm$4} \\
& $\tau$=.5 & 32.5{\tiny$\pm$1.2} & 1121{\tiny$\pm$10} & 83.2{\tiny$\pm$1.3} & 366{\tiny$\pm$9} & 82.9{\tiny$\pm$1.0} & 771{\tiny$\pm$21} & 86.1{\tiny$\pm$.3} & 553{\tiny$\pm$5} & 57.2{\tiny$\pm$1.5} & 1186{\tiny$\pm$35} & 79.7{\tiny$\pm$1.1} & 1494{\tiny$\pm$24} \\
& $\tau$=.7 & 39.5{\tiny$\pm$.5} & 627{\tiny$\pm$16} & 83.4{\tiny$\pm$.1} & 354{\tiny$\pm$8} & 79.9{\tiny$\pm$.4} & 701{\tiny$\pm$31} & 83.3{\tiny$\pm$.7} & 493{\tiny$\pm$13} & 55.6{\tiny$\pm$.9} & 1156{\tiny$\pm$16} & 78.7{\tiny$\pm$1.3} & 1228{\tiny$\pm$21} \\
& $\tau$=.9 & 59.2{\tiny$\pm$.3} & 85{\tiny$\pm$2} & 79.6{\tiny$\pm$.7} & 175{\tiny$\pm$10} & 79.3{\tiny$\pm$.5} & 686{\tiny$\pm$12} & 74.6{\tiny$\pm$.3} & 353{\tiny$\pm$14} & 54.3{\tiny$\pm$.5} & 1047{\tiny$\pm$11} & 69.9{\tiny$\pm$1.3} & 1074{\tiny$\pm$58} \\
\bottomrule
\end{tabular}%
}
\end{table}

\newpage
\section{Out-of-distribution generalization}
\label{app:ood}

\begin{figure}[H]
\centering
\includegraphics[width=\textwidth]{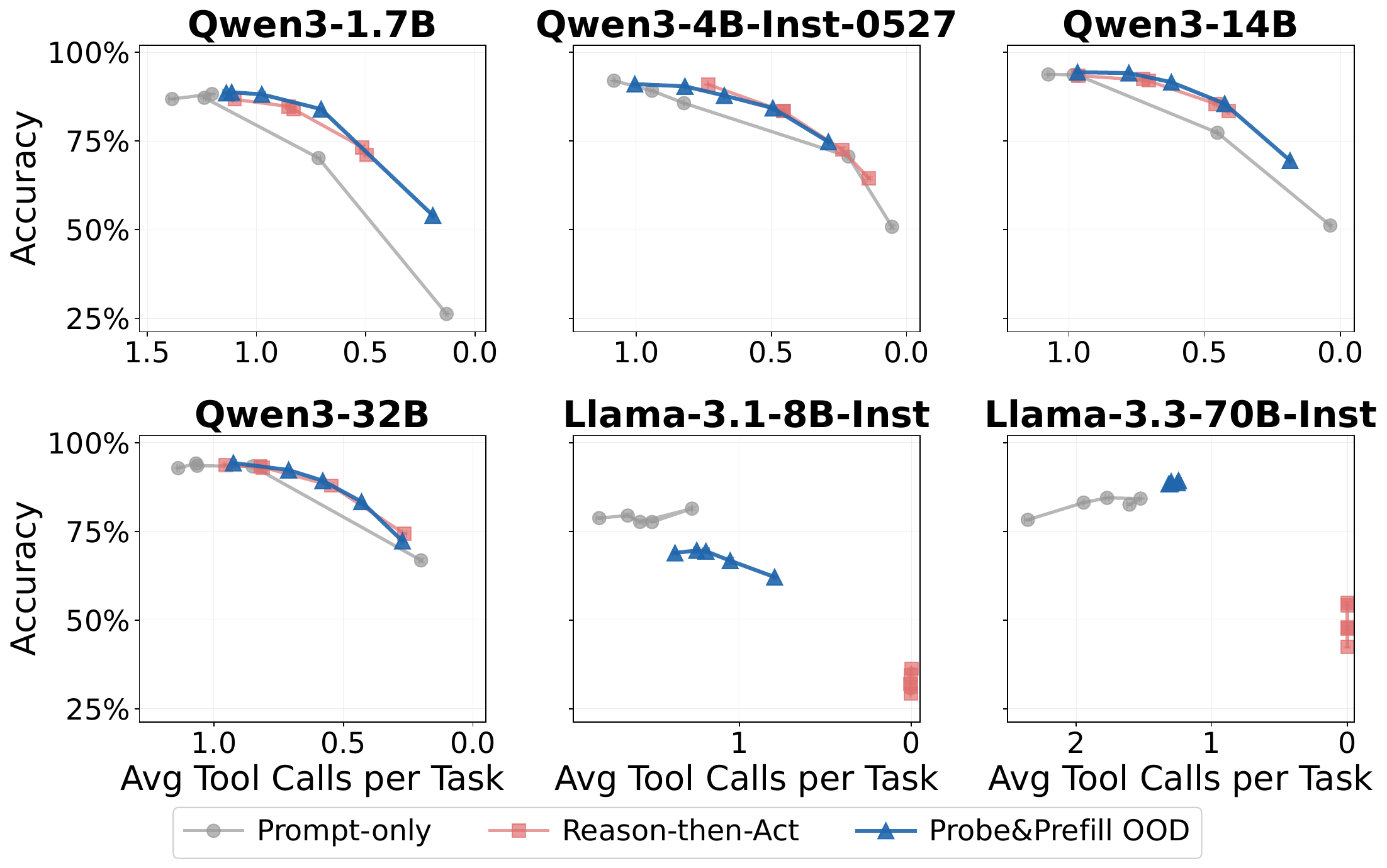}
\caption{OOD generalization: in-distribution probe (green) vs.\ OOD probe trained on held-out environments (blue). Gray and red lines show Prompt-only and Reason-then-Act baselines. The OOD probe closely tracks the in-distribution probe across all models.}
\label{fig:abl_ood}
\end{figure}

To test whether the probe generalizes beyond its training environments, we conduct within-category leave-two-out experiments: for each category, we train the probe on 3 of the 5 environments and evaluate on all 5. Figure~\ref{fig:abl_ood} compares the OOD probe (blue) against the in-distribution probe (green). The OOD probe achieves comparable accuracy--efficiency tradeoffs across all models, confirming that the probe learns general signals rather than environment-specific shortcuts.

\newpage
\section{Ablation studies}
\label{app:ablations}

\subsection{Soft vs.\ hard prefill}
\label{app:ablation_prefill}

\begin{figure}[H]
\centering
\includegraphics[width=\textwidth]{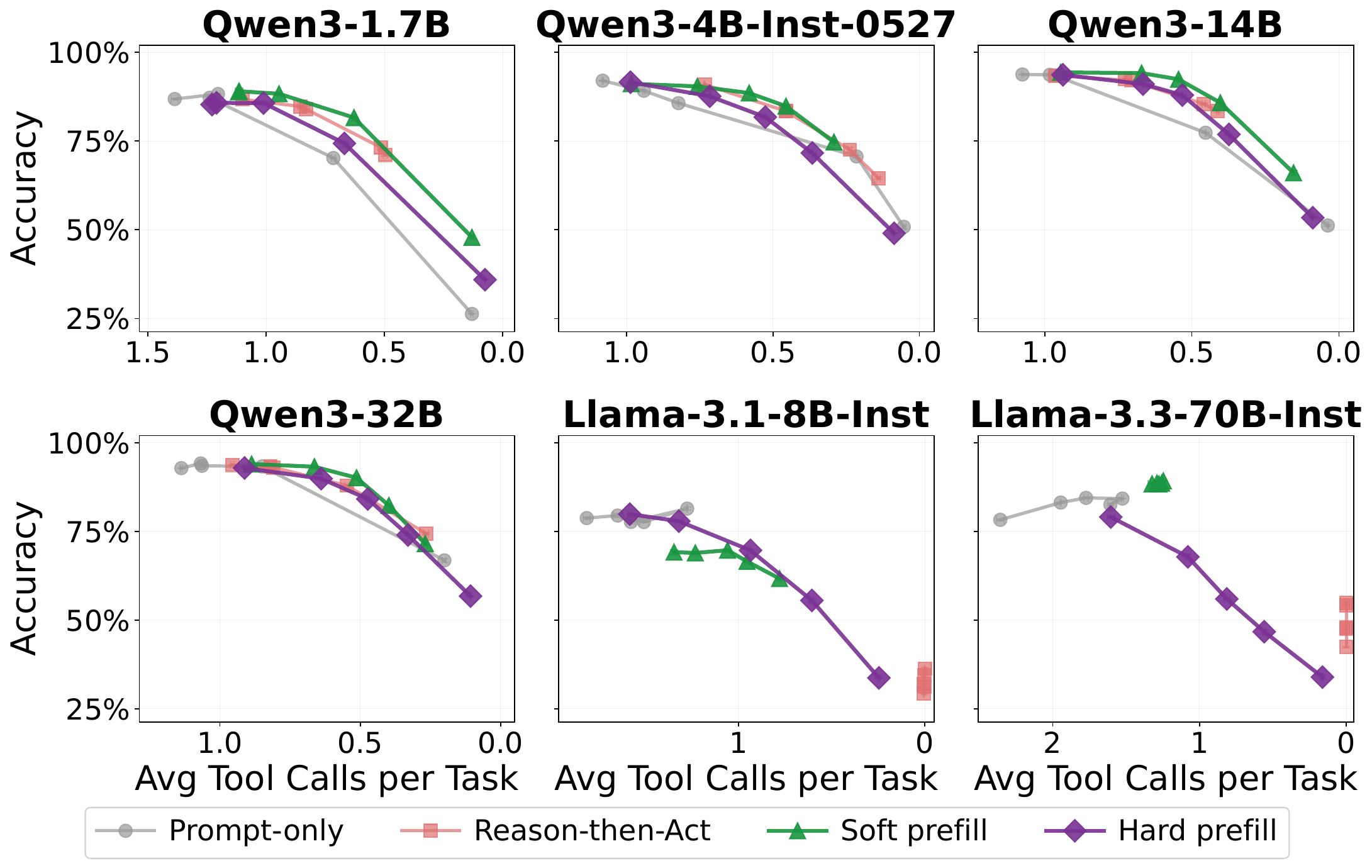}
\caption{Soft prefill (green) vs.\ hard prefill (purple). Hard prefill forces the output format, while soft prefill allows the model to override.}
\label{fig:abl_prefill}
\end{figure}

Soft prefill injects a natural language steering sentence that the model may override. Hard prefill forces the output format (\texttt{\textbackslash boxed\{} for direct answers, \texttt{\{"name":} for tool calls), leaving no room for deviation. Figure~\ref{fig:abl_prefill} visualizes the tradeoff curves. On Qwen models, soft prefill generally achieves higher accuracy than hard at matched tool-call levels, because the model can override incorrect predictions. On Llama-3.1-8B, hard prefill achieves better accuracy at low thresholds (79.9\% vs.\ 69.2\% at $\tau$=0.1) because the soft steering sentence is frequently ignored. On Llama-3.3-70B, soft prefill is remarkably stable (88.4--89.2\% across all thresholds) because the model largely ignores the steering, while hard prefill provides a wide range (79.0\% at $\tau$=0.1 to 33.9\% at $\tau$=0.9), confirming that forcing the output format is the only way to control this model's tool-call behavior. Table~\ref{tab:abl_prefill} compares both modes across thresholds.

\vspace{-10pt}
\begin{table}[H]
\caption{Soft vs.\ hard prefill ($T$=2.0). Each cell: Acc (\%) / total tool calls.}
\label{tab:abl_prefill}
\centering
\small
\setlength{\tabcolsep}{3pt}
\begin{tabular}{llccccc}
\toprule
Model & Mode & $\tau$=0.1 & $\tau$=0.3 & $\tau$=0.5 & $\tau$=0.7 & $\tau$=0.9 \\
\midrule
\multirow{2}{*}{Qwen3-1.7B}
  & Soft & 88.8/2512 & 89.0/2507 & 88.3/2128 & 81.6/1415 & 47.9/293 \\
  & Hard & 85.3/2765 & 85.7/2723 & 85.7/2275 & 74.3/1506 & 35.9/169 \\
\midrule
\multirow{2}{*}{Qwen3-4B}
  & Soft & 91.1/2216 & 90.4/1707 & 88.5/1309 & 84.8/1026 & 74.7/657 \\
  & Hard & 91.6/2222 & 87.6/1612 & 81.7/1185 & 71.6/824  & 49.0/195 \\
\midrule
\multirow{2}{*}{Qwen3-14B}
  & Soft & 94.3/2128 & 94.1/1509 & 92.4/1227 & 85.8/907  & 66.0/347 \\
  & Hard & 93.6/2111 & 91.1/1498 & 87.9/1198 & 76.9/841  & 53.4/200 \\
\midrule
\multirow{2}{*}{Qwen3-32B}
  & Soft & 94.0/1996 & 93.2/1493 & 90.1/1155 & 82.3/896  & 71.5/604 \\
  & Hard & 92.9/2053 & 89.9/1439 & 84.1/1064 & 74.0/741  & 56.8/241 \\
\midrule
\multirow{2}{*}{Llama-3.1-8B}
  & Soft & 69.2/3027 & 68.9/2770 & 69.7/2381 & 66.5/2146 & 61.7/1753 \\
  & Hard & 79.9/3561 & 77.9/2969 & 69.6/2105 & 55.5/1363 & 33.7/554 \\
\midrule
\multirow{2}{*}{Llama-3.3-70B}
  & Soft & 88.4/2976 & 88.6/2902 & 88.3/2871 & 88.6/2828 & 89.2/2804 \\
  & Hard & 79.0/3609 & 67.8/2427 & 56.0/1830 & 46.7/1258 & 33.9/366 \\
\bottomrule
\end{tabular}
\end{table}

\subsection{Temperature scaling}
\label{app:ablation_temp}

The probe outputs a logit $z = \mathbf{w}^\top \mathbf{x} + b$, which is converted to a probability via $p = \sigma(z / T)$ before thresholding. Higher $T$ flattens the distribution, making the probe more conservative about predicting tool necessity. Table~\ref{tab:abl_temp} compares $T \in \{1.0, 2.0, 3.0\}$.

\begin{figure}[H]
\centering
\includegraphics[width=\textwidth]{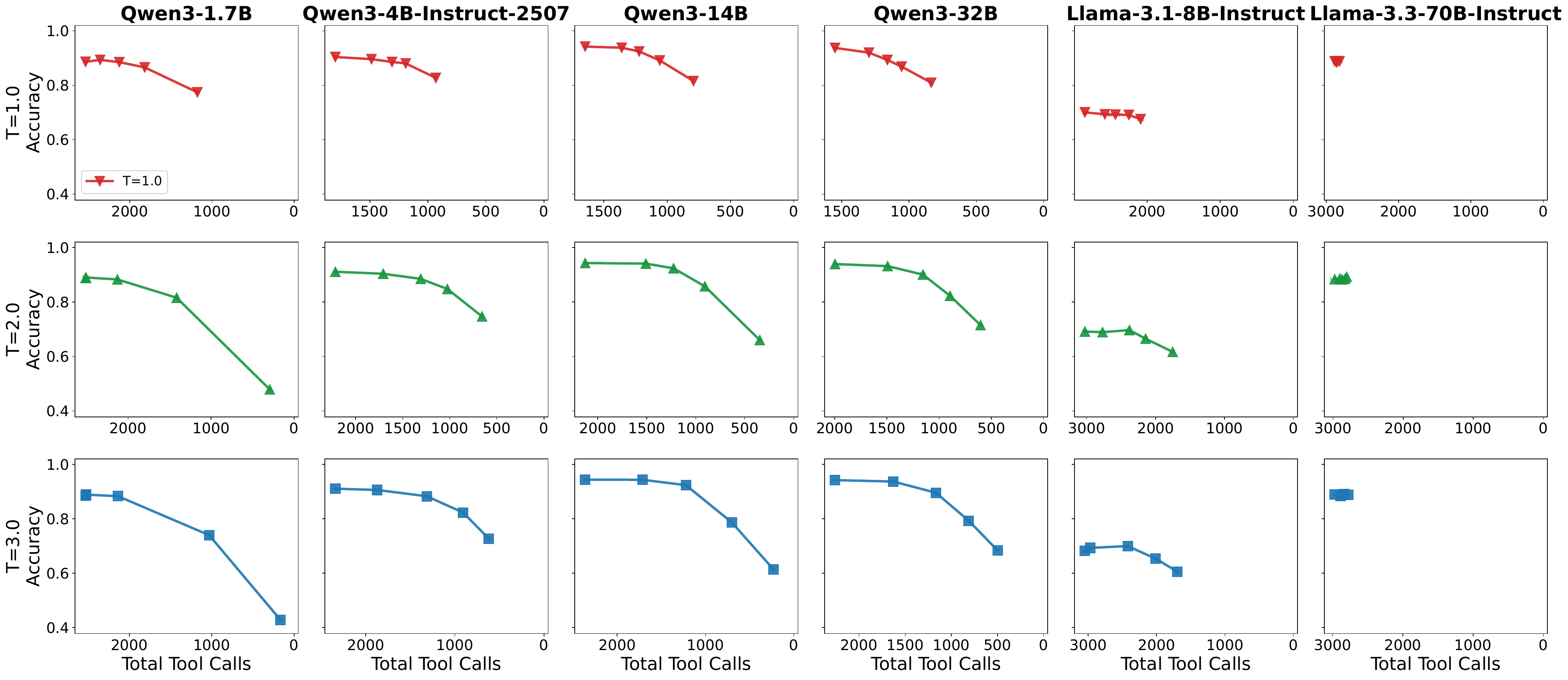}
\caption{Temperature scaling: $T$=1.0 (red), $T$=2.0 (green, default), $T$=3.0 (blue). Higher temperature flattens the probe's confidence distribution.}
\label{fig:abl_temp}
\end{figure}

Figure~\ref{fig:abl_temp} visualizes the tradeoff curves. At $T$=1.0, the probe is sharp: low thresholds already reduce many tool calls, providing a wider operating range. At $T$=3.0, the probe is diffuse, offering finer control in the middle range. $T$=2.0 provides a good balance across models. The choice of temperature does not qualitatively change the finding that \method{} outperforms prompt baselines.

\begin{table}[H]
\caption{Temperature scaling (soft prefill). Each cell: Acc (\%) / total tool calls.}
\label{tab:abl_temp}
\centering
\small
\setlength{\tabcolsep}{3pt}
\begin{tabular}{llccccc}
\toprule
Model & $T$ & $\tau$=0.1 & $\tau$=0.3 & $\tau$=0.5 & $\tau$=0.7 & $\tau$=0.9 \\
\midrule
\multirow{3}{*}{Qwen3-1.7B}
  & 1.0 & 88.6/2539 & 89.3/2361 & 88.5/2128 & 86.6/1819 & 77.4/1178 \\
  & 2.0 & 88.8/2512 & 89.0/2507 & 88.3/2128 & 81.6/1415 & 47.9/293 \\
  & 3.0 & 88.5/2532 & 88.9/2526 & 88.3/2139 & 74.0/1031 & 42.8/165 \\
\midrule
\multirow{3}{*}{Qwen3-4B}
  & 1.0 & 90.4/1797 & 89.6/1486 & 88.6/1309 & 88.0/1191 & 82.7/931 \\
  & 2.0 & 91.1/2216 & 90.4/1707 & 88.5/1309 & 84.8/1026 & 74.7/657 \\
  & 3.0 & 91.1/2342 & 90.6/1872 & 88.3/1315 & 82.3/907  & 72.7/620 \\
\midrule
\multirow{3}{*}{Qwen3-14B}
  & 1.0 & 94.2/1648 & 93.8/1359 & 92.4/1221 & 89.1/1058 & 81.6/792 \\
  & 2.0 & 94.3/2128 & 94.1/1509 & 92.4/1227 & 85.8/907  & 66.0/347 \\
  & 3.0 & 94.4/2362 & 94.4/1714 & 92.4/1221 & 78.7/701  & 61.4/230 \\
\midrule
\multirow{3}{*}{Qwen3-32B}
  & 1.0 & 93.7/1550 & 92.0/1298 & 89.3/1161 & 86.9/1055 & 80.9/835 \\
  & 2.0 & 94.0/1996 & 93.2/1493 & 90.1/1155 & 82.3/896  & 71.5/604 \\
  & 3.0 & 94.2/2260 & 93.7/1630 & 89.6/1168 & 79.3/812  & 68.4/498 \\
\midrule
\multirow{3}{*}{Llama-3.1-8B}
  & 1.0 & 70.1/2852 & 69.3/2578 & 69.2/2434 & 69.1/2250 & 67.5/2090 \\
  & 2.0 & 69.2/3027 & 68.9/2770 & 69.7/2381 & 66.5/2146 & 61.7/1753 \\
  & 3.0 & 68.2/3048 & 69.3/2973 & 70.0/2422 & 65.3/2014 & 60.6/1696 \\
\midrule
\multirow{3}{*}{Llama-3.3-70B}
  & 1.0 & 88.8/2882 & 88.6/2860 & 88.9/2841 & 88.5/2854 & 88.8/2815 \\
  & 2.0 & 88.4/2976 & 88.6/2902 & 88.3/2871 & 88.6/2828 & 89.2/2804 \\
  & 3.0 & 88.9/2971 & 88.4/2890 & 88.8/2853 & 88.8/2780 & 89.0/2835 \\
\bottomrule
\end{tabular}
\end{table}

\subsection{Layer selection}
\label{app:ablation_layer}

We compare three probe configurations: all layers concatenated, middle layer only, and last layer only. Table~\ref{tab:abl_layer} reports probe test AUROC and accuracy. All-layer concatenation consistently performs best, confirming that tool-necessity information is distributed across the network. Single-layer probes remain competitive, with the mid-layer probe slightly outperforming the last-layer probe on most models, suggesting the signal emerges early and persists through the network.

\begin{table}[H]
\caption{Layer selection: probe test AUROC and accuracy.}
\label{tab:abl_layer}
\centering
\small
\begin{tabular}{llcc}
\toprule
Model & Layers & AUROC & Accuracy \\
\midrule
\multirow{3}{*}{Qwen3-1.7B}
  & All (29 layers) & 0.894 & 0.847 \\
  & Mid (layer 14)  & 0.835 & 0.795 \\
  & Last (layer 28) & 0.863 & 0.796 \\
\midrule
\multirow{3}{*}{Qwen3-4B}
  & All (37 layers) & 0.948 & 0.877 \\
  & Mid (layer 18)  & 0.916 & 0.860 \\
  & Last (layer 36) & 0.893 & 0.805 \\
\midrule
\multirow{3}{*}{Qwen3-14B}
  & All (41 layers) & 0.957 & 0.892 \\
  & Mid (layer 20)  & 0.920 & 0.851 \\
  & Last (layer 40) & 0.929 & 0.865 \\
\midrule
\multirow{3}{*}{Qwen3-32B}
  & All (65 layers) & 0.952 & 0.885 \\
  & Mid (layer 32)  & 0.921 & 0.844 \\
  & Last (layer 64) & 0.916 & 0.835 \\
\midrule
\multirow{3}{*}{Llama-3.1-8B}
  & All (33 layers) & 0.927 & 0.849 \\
  & Mid (layer 16)  & 0.894 & 0.805 \\
  & Last (layer 32) & 0.880 & 0.795 \\
\midrule
\multirow{3}{*}{Llama-3.3-70B}
  & All (81 layers) & 0.936 & 0.872 \\
  & Mid (layer 40)  & 0.912 & 0.839 \\
  & Last (layer 80) & 0.908 & 0.828 \\
\bottomrule
\end{tabular}
\end{table}

\subsection{Data efficiency}
\label{app:ablation_data}

We subsample the 900-example training set to $\{10\%, 25\%, 50\%, 75\%, 100\%\}$ and retrain the probe. Table~\ref{tab:abl_data} shows that even with only 90 labeled examples (10\%), the probe achieves AUROC above 0.81 on all tested models. Performance improves steadily with more data, but the marginal gain diminishes beyond 50\%, suggesting the signal is easy to extract with minimal supervision.

\begin{table}[H]
\caption{Data efficiency: probe test AUROC at varying training fractions.}
\label{tab:abl_data}
\centering
\small
\begin{tabular}{lccccc}
\toprule
Model & 10\% & 25\% & 50\% & 75\% & 100\% \\
\midrule
Qwen3-1.7B      & 0.813 & 0.872 & 0.880 & 0.888 & 0.894 \\
Qwen3-4B        & 0.894 & 0.926 & 0.936 & 0.943 & 0.948 \\
Qwen3-14B       & 0.882 & 0.929 & 0.937 & 0.949 & 0.957 \\
Qwen3-32B       & 0.887 & 0.926 & 0.942 & 0.948 & 0.952 \\
Llama-3.1-8B    & 0.823 & 0.885 & 0.914 & 0.920 & 0.927 \\
Llama-3.3-70B   & 0.826 & 0.908 & 0.916 & 0.931 & 0.936 \\
\bottomrule
\end{tabular}
\end{table}

\subsection{Regularization strength}
\label{app:ablation_reg}

We sweep the L2 regularization parameter $\lambda \in \{1, 10, 100, 1000, 10000, 100000\}$ (sklearn $C = 1/\lambda$). Table~\ref{tab:abl_reg} shows that probe performance is stable across four orders of magnitude ($\lambda$=10 to 10000), with a modest decline at very low regularization ($\lambda$=1) and very high regularization ($\lambda$=100000). The default $\lambda$=10000 used throughout the paper is near-optimal across all models.

\begin{table}[H]
\caption{Regularization strength: probe test AUROC.}
\label{tab:abl_reg}
\centering
\small
\begin{tabular}{lcccccc}
\toprule
Model & $\lambda$=1 & 10 & 100 & 1000 & 10000 & 100000 \\
\midrule
Qwen3-1.7B & 0.863 & 0.873 & 0.882 & 0.892 & 0.894 & 0.879 \\
Qwen3-4B   & 0.936 & 0.940 & 0.944 & 0.949 & 0.948 & 0.935 \\
Qwen3-14B  & 0.947 & 0.951 & 0.952 & 0.955 & 0.957 & 0.947 \\
Qwen3-32B     & 0.935 & 0.940 & 0.944 & 0.948 & 0.952 & 0.947 \\
Llama-3.1-8B  & 0.907 & 0.916 & 0.922 & 0.928 & 0.927 & 0.905 \\
Llama-3.3-70B & 0.915 & 0.921 & 0.925 & 0.929 & 0.936 & 0.934 \\
\bottomrule
\end{tabular}
\end{table}

\newpage
\section{Inference overhead}
\label{app:overhead}

Table~\ref{tab:overhead} reports the computational overhead of \method{} at inference time. The probe requires two operations beyond the standard generation pipeline: (1) extracting the hidden state at the last token position from the prefill forward pass, and (2) applying the linear probe (standardization + dot product + sigmoid). The forward pass itself is \emph{not} additional cost, as it is already required to build the KV cache before autoregressive generation begins. The only overhead is reading the hidden states from this existing forward pass and running the probe.

Across all six models, the total additional latency is under 0.7~ms, compared to a typical prefill forward pass of 10--100~ms and per-token generation of 5--50~ms. This represents less than 1\% overhead, making \method{} essentially free at inference time.

\begin{table}[H]
\caption{Inference overhead of \method{}. The forward pass is shared with standard generation (no additional cost). The only overhead is hidden state extraction and the linear probe.}
\label{tab:overhead}
\centering
\small
\begin{tabular}{lcccc}
\toprule
Model & Layers & Hidden dim & Probe dim & Overhead (ms) \\
\midrule
Qwen3-1.7B      & 29 & 2,048 & 59,392  & 0.38 \\
Qwen3-4B        & 37 & 2,560 & 94,720  & 0.38 \\
Llama-3.1-8B    & 33 & 4,096 & 135,168 & 0.35 \\
Qwen3-14B       & 41 & 5,120 & 209,920 & 0.40 \\
Qwen3-32B       & 65 & 5,120 & 332,800 & 0.57 \\
Llama-3.3-70B   & 81 & 8,192 & 663,552 & 0.69 \\
\bottomrule
\end{tabular}
\end{table}

\newpage
\section{Generalization to agentic search (Search-o1 benchmarks)}
\label{app:search_o1}

To evaluate whether \method{} generalizes beyond \benchmark{}, we apply it to six open-domain QA benchmarks from the Search-o1 framework~\citep{li2025search}: NQ, TriviaQA, HotpotQA, 2WikiMultihopQA, Bamboogle, and MuSiQue. These benchmarks cover single-hop factual QA (NQ, TriviaQA), two-hop reasoning (HotpotQA, 2WikiMultihopQA), and complex multi-hop reasoning (MuSiQue, Bamboogle). We use Qwen3-4B-Instruct for all the evaluations.

\paragraph{Setup.}
We follow the same experimental design as \benchmark{}: 5 Prompt-only modes and 5 Reason-then-Act modes as baselines, plus \method{} with threshold sweep. We train the probes on a 50/50 split of each dataset. The probe trains in seconds on CPU.

\paragraph{Summary.}
Table~\ref{tab:search_o1_summary} compares the best baseline (highest accuracy among all Prompt-only and Reason-then-Act settings) against the best \method{} operating point for each dataset. On 4 of 6 datasets, \method{} achieves comparable or better accuracy while reducing search calls more than the best baseline. On TriviaQA, \method{} achieves 69.2\% accuracy (vs.\ best baseline 68.8\%) with 20\% fewer searches compared to the baseline's 16\%. On HotpotQA and Bamboogle, \method{} \emph{exceeds} all baselines in accuracy while using 50--54\% fewer searches. MuSiQue is the exception: this 3--4 hop dataset requires nearly all questions to be searched, and the best baseline (PO Force, $-$56\%) achieves stronger reduction than \method{} ($-$48\%).

\begin{table}[H]
\caption{Search-o1 generalization summary. All on 50\% held-out test split. $\Delta$TC = search call reduction relative to Default. \textbf{Bold} = probe beats all baselines.}
\label{tab:search_o1_summary}
\centering
\small
\setlength{\tabcolsep}{3pt}
\begin{tabular}{l c cc ccc ccc}
\toprule
& & \multicolumn{2}{c}{Default} & \multicolumn{3}{c}{Best Baseline} & \multicolumn{3}{c}{\method{}} \\
\cmidrule(lr){3-4} \cmidrule(lr){5-7} \cmidrule(lr){8-10}
Dataset & $N$ & Acc & TC & Acc & TC & $\Delta$TC & Acc & TC & $\Delta$TC \\
\midrule
NQ          & 250 & 44.8 & 263  & \textbf{44.8} & 252  & $-$4\%   & 42.0 & 248  & \textbf{$-$6\%} \\
TriviaQA    & 250 & 69.6 & 344  & 68.8 & 289  & $-$16\%  & \textbf{69.2} & 275  & \textbf{$-$20\%} \\
HotpotQA    & 250 & 26.0 & 802  & 27.2 & 713  & $-$11\%  & \textbf{28.8} & 404  & \textbf{$-$50\%} \\
2Wiki       & 250 & 36.4 & 670  & 38.4 & 561  & $-$16\%  & \textbf{39.2} & 297  & \textbf{$-$56\%} \\
Bamboogle   &  63 & 25.4 & 188  & 33.3 & 145  & $-$23\%  & \textbf{34.9} & 87   & \textbf{$-$54\%} \\
MuSiQue     & 250 & 19.6 & 829  & \textbf{20.4} & 362  & \textbf{$-$56\%}  & \textbf{20.4} & 431  & $-$48\% \\
\bottomrule
\end{tabular}
\end{table}

\paragraph{Probe quality.}
Table~\ref{tab:search_o1_auroc} reports probe AUROC on the held-out test split. We compare two probes: (1) \emph{\benchmark{} transfer}, the probe trained on \benchmark{} and applied directly without any retraining, and (2) \emph{in-domain}, a probe trained on 250 in-domain examples from each dataset. The \benchmark{} transfer probe achieves AUROC 0.67--0.80, confirming that the tool-necessity signal learned on our controlled benchmark partially transfers to real-world QA tasks. The in-domain probe (trained in seconds) achieves AUROC 0.64--0.84, with the strongest signal on TriviaQA (0.84) and 2WikiMultihopQA (0.80).

\begin{table}[H]
\caption{Probe AUROC on Search-o1 benchmarks (test split).}
\label{tab:search_o1_auroc}
\centering
\small
\begin{tabular}{lccc}
\toprule
Dataset & $N$ & \benchmark{} transfer & In-domain \\
\midrule
NQ          & 250 & 0.731 & 0.746 \\
TriviaQA    & 250 & 0.787 & 0.835 \\
HotpotQA    & 250 & 0.742 & 0.798 \\
2Wiki       & 250 & 0.803 & 0.803 \\
Bamboogle   &  63 & 0.675 & 0.802 \\
MuSiQue     & 250 & 0.695 & 0.640 \\
\bottomrule
\end{tabular}
\end{table}

\paragraph{Full results.}
Tables~\ref{tab:search_o1_full_nr}--\ref{tab:search_o1_full_probe} report complete results across all Prompt-only, Reason-then-Act, and \method{} settings. Acc = accuracy (substring match), EM = exact match, TC = total search calls.

\begin{table}[H]
\caption{Full results on Search-o1 QA benchmarks (test split only). Acc = accuracy (\%), EM = exact match (\%), TC = total searches. Baselines evaluated on 250 test items (63 for Bamboogle); \method{} evaluated on the probe's held-out test split.}
\label{tab:search_o1_full_nr}
\label{tab:search_o1_full_probe}
\centering
\resizebox{\textwidth}{!}{%
\small
\setlength{\tabcolsep}{2.5pt}
\renewcommand{\arraystretch}{1.3}
\begin{tabular}{ll ccc ccc ccc ccc ccc ccc}
\toprule
& & \multicolumn{3}{c}{NQ} & \multicolumn{3}{c}{TriviaQA} & \multicolumn{3}{c}{HotpotQA} & \multicolumn{3}{c}{2Wiki} & \multicolumn{3}{c}{Bamboogle} & \multicolumn{3}{c}{MuSiQue} \\
\cmidrule(lr){3-5} \cmidrule(lr){6-8} \cmidrule(lr){9-11} \cmidrule(lr){12-14} \cmidrule(lr){15-17} \cmidrule(lr){18-20}
& & Acc & EM & TC & Acc & EM & TC & Acc & EM & TC & Acc & EM & TC & Acc & EM & TC & Acc & EM & TC \\
\midrule
\multirow{5}{*}{\rotatebox{90}{Prompt-only}}
& F & 44.8 & 30.4 & 252 & 68.8 & 58.8 & 289 & 26.0 & 22.8 & 367 & 36.0 & 30.8 & 312 & 30.2 & 28.6 & 78 & 20.4 & 18.4 & 362 \\
& D & 44.8 & 31.6 & 263 & 69.6 & 60.4 & 344 & 26.0 & 24.0 & 802 & 36.4 & 30.8 & 670 & 25.4 & 23.8 & 188 & 19.6 & 16.8 & 829 \\
& N & 42.8 & 30.4 & 377 & 67.6 & 58.0 & 465 & 26.4 & 24.0 & 726 & 41.2 & 35.2 & 706 & 23.8 & 22.2 & 180 & 20.0 & 17.6 & 695 \\
& S & 42.0 & 30.0 & 353 & 65.6 & 58.0 & 347 & 24.8 & 21.6 & 452 & 34.8 & 31.2 & 477 & 23.8 & 20.6 & 105 & 12.8 & 10.4 & 415 \\
& X & 35.6 & 24.8 & 170 & 53.2 & 46.8 & 137 & 19.2 & 17.2 & 166 & 28.8 & 26.8 & 159 & 23.8 & 22.2 & 58 & 8.8 & 6.8 & 157 \\
\midrule
\multirow{5}{*}{\rotatebox{90}{Reason-then-Act}}
& F & 45.6 & 31.6 & 272 & 68.8 & 58.8 & 363 & 27.2 & 24.8 & 713 & 38.4 & 32.0 & 561 & 33.3 & 31.7 & 145 & 20.0 & 17.2 & 692 \\
& D & 42.0 & 28.8 & 369 & 69.2 & 60.0 & 444 & 23.6 & 21.2 & 842 & 38.0 & 31.6 & 785 & 27.0 & 25.4 & 195 & 19.2 & 17.2 & 899 \\
& N & 42.0 & 29.2 & 314 & 67.2 & 58.0 & 366 & 24.0 & 20.8 & 669 & 37.6 & 31.6 & 688 & 23.8 & 22.2 & 175 & 17.2 & 15.6 & 690 \\
& S & 40.8 & 29.6 & 363 & 65.6 & 56.4 & 366 & 23.2 & 21.2 & 631 & 33.6 & 30.0 & 592 & 25.4 & 23.8 & 144 & 16.0 & 12.8 & 568 \\
& X & 39.6 & 27.2 & 255 & 60.8 & 52.0 & 247 & 21.6 & 19.6 & 301 & 28.0 & 25.6 & 295 & 23.8 & 20.6 & 80 & 10.0 & 8.4 & 263 \\
\midrule
\multirow{5}{*}{\rotatebox{90}{\method{}}}
& $\tau$=.1 & 41.6 & 29.2 & 263 & 69.2 & 60.4 & 275 & 28.8 & 25.6 & 404 & 38.4 & 33.2 & 337 & 33.3 & 31.7 & 97 & 18.8 & 17.2 & 471 \\
& $\tau$=.3 & 42.0 & 28.8 & 248 & 68.0 & 60.0 & 211 & 28.4 & 26.8 & 379 & 39.2 & 32.4 & 297 & 27.0 & 25.4 & 82 & 18.4 & 16.0 & 433 \\
& $\tau$=.5 & 41.2 & 27.6 & 233 & 60.4 & 53.6 & 181 & 25.6 & 23.2 & 372 & 36.4 & 30.4 & 266 & 34.9 & 34.9 & 87 & 20.4 & 18.0 & 431 \\
& $\tau$=.7 & 40.8 & 27.6 & 194 & 56.4 & 50.0 & 117 & 26.4 & 24.4 & 274 & 33.6 & 28.8 & 175 & 31.7 & 27.0 & 80 & 20.0 & 16.8 & 397 \\
& $\tau$=.9 & 33.6 & 22.4 & 92 & 46.4 & 41.2 & 33 & 20.0 & 18.8 & 175 & 32.0 & 25.6 & 109 & 33.3 & 30.2 & 37 & 13.2 & 11.6 & 285 \\
\bottomrule
\end{tabular}%
}
\end{table}

\newpage
\section{Additional baseline: Compare \method{} with Supervised Fine-Tuning (SFT)}
\label{app:sft}

To provide a stronger baseline, we compare \method{} against supervised fine-tuning (SFT) that directly modifies model weights to learn tool-call decisions. We note that this comparison is inherently asymmetric: SFT requires full fine-tuning on multiple GPUs for hours, while \method{} trains a linear probe in seconds on CPU with no weight modification.

\subsection{Original training setting: 900 examples}
\label{app:sft_original}

\paragraph{Training data collection.}
We construct SFT training data from the 900 single-hop training tasks using the same binary labels as the probe. For each task, we first evaluate the model without tool access. If the model answers correctly (tool unnecessary, $y{=}0$), we use that direct-answer trajectory as the training target. If the model fails (tool necessary, $y{=}1$), we run the model with the Default prompt and tools available, collecting the full multi-round trajectory (tool call, tool response, and final answer). This gives the model examples of both when to answer directly and when to call tools.

\paragraph{Training setup.}
We perform full-parameter fine-tuning using the HuggingFace Trainer with gradient checkpointing across 4 GPUs. We train for 2 epochs with learning rate $10^{-5}$ and mask all non-assistant tokens (system prompt, user messages, tool responses) so the model only learns from its own response tokens at each turn. We evaluate on Qwen3-1.7B, Qwen3-4B-Instruct, and Llama-3.1-8B.

\paragraph{Results.}
Table~\ref{tab:sft_results} compares SFT against the Default baseline and \method{} ($\tau$=0.5). SFT improves accuracy by 2--3\% across all three models, which is expected since the training data provides correct trajectories directly. However, SFT does not consistently reduce tool calls: on Qwen3-4B, tool calls slightly increase, and on Llama-3.1-8B they also increase. Only on Qwen3-1.7B does SFT achieve meaningful reduction ($-$18\%). In contrast, \method{} at $\tau$=0.5 reduces tool calls by 21--38\% on the all models.

\begin{table}[H]
\caption{SFT (full fine-tuning) vs.\ \method{} ($\tau$=0.5). SFT improves accuracy but does not reliably reduce tool calls. \method{} achieves larger TC reductions with no weight modification. Mean over 3 runs.}
\label{tab:sft_results}
\centering
\small
\setlength{\tabcolsep}{4pt}
\begin{tabular}{llcc}
\toprule
Model & Method & Acc (\%) & Total TC \\
\midrule
\multirow{3}{*}{Qwen3-1.7B}
  & Default baseline       & 88.2 & 2709 \\
  & SFT                    & \textbf{91.0} & 2211 \\
  & \method{} ($\tau$=0.5) & 88.3 & \textbf{2128} \\
\midrule
\multirow{3}{*}{Qwen3-4B-Inst.}
  & Default baseline       & 89.2 & 2118 \\
  & SFT                    & \textbf{91.3} & 2140 \\
  & \method{} ($\tau$=0.5) & 88.5 & \textbf{1309} \\
\midrule
\multirow{3}{*}{Llama-3.1-8B}
  & Default baseline       & 79.5 & 3708 \\
  & SFT                    & \textbf{81.7} & 3777 \\
  & \method{} ($\tau$=0.5) & 69.7 & \textbf{2381} \\
\bottomrule
\end{tabular}
\end{table}

\paragraph{Discussion.}
SFT learns to produce better answers overall but does not learn the tool-call \emph{decision boundary} effectively. In contrast, \method{} works consistently across all model sizes (1.7B to 70B), requires only seconds of CPU training, and provides a smooth accuracy--efficiency tradeoff via the threshold $\tau$. These results highlight that \method{} offers an effective, nearly zero-cost solution compared to the substantially more expensive SFT alternative.

\subsection{Larger training setting: 6,000 examples}
\label{app:scaled_sft}

To test whether the limited call reduction of the 900-example SFT baseline is due to training-set size, we extend supervised training to 6{,}000 examples on Qwen3-4B-Instruct and evaluate both full-parameter SFT and LoRA with rank $r=16$. The supervision remains model-specific: direct-answer and tool-use targets are determined by the model's measured no-tool outcomes, rather than by a separate teacher model. We evaluate on the same 2{,}250 single-hop test tasks over three runs.

\begin{table}[H]
\caption{Scaled-data SFT and LoRA on Qwen3-4B-Instruct. Accuracy is reported as mean $\pm$ standard deviation over three runs; calls/task is the mean number of tool calls per test task. The probe uses its original 900-example training set.}
\label{tab:scaled_sft}
\centering
\small
\begin{tabular}{lrr}
\toprule
Method & Acc (\%) & Calls/task \\
\midrule
Default baseline & $89.2\pm0.2$ & 0.941 \\
SFT, 900 examples & $91.3\pm0.1$ & 0.951 \\
SFT, 6{,}000 examples & $90.0\pm0.2$ & 0.998 \\
LoRA, 6{,}000 examples & $89.6\pm0.3$ & 0.937 \\
\method{}, $\tau=0.1$ & $91.1\pm0.3$ & 0.985 \\
\method{}, $\tau=0.5$ & $88.5\pm0.2$ & \textbf{0.582} \\
\bottomrule
\end{tabular}
\end{table}

Table~\ref{tab:scaled_sft} shows that scaling this training recipe does not yield substantial tool-call savings: full SFT increases calls to 0.998 per task, while LoRA remains close to Default at 0.937. In contrast, \method{} at $\tau=0.5$ reduces calls by approximately 38\% relative to Default, with accuracy decreasing from 89.2\% to 88.5\%. These results characterize the evaluated training recipes and do not rule out stronger objectives or alternative fine-tuning configurations for calibrated tool use.

\newpage
\section{Additional baselines: Compare \method{} with Confidence-Based Tool-Use Controllers}
\label{app:confidence_controllers}

We compare the hidden-state probe with three confidence-based controllers on Qwen3-4B-Instruct. All controllers drive the same soft-prefill mechanism and evaluation loop as \method{}, so the comparison changes the decision signal while preserving the downstream intervention. We evaluate on 2{,}250 single-hop test tasks and report means over three runs.

\paragraph{Decision signals.}
\textbf{Verbalized confidence} asks the model for a score from 0 to 100 indicating its confidence in answering without tools; we convert confidence $c$ to a necessity score $1-c/100$. \textbf{First-token likelihood} uses the model's probability of emitting a tool-call token at the first generated position. \textbf{Self-ask} asks whether the model can solve the task without tools and uses the probability assigned to ``No,'' normalized over the ``Yes'' and ``No'' alternatives. All scores are oriented so that larger values favor tool use.

\paragraph{Thresholds and comparison protocol.}
For each confidence baseline, we set the decision cutoff to the empirical score quantiles $q\in\{0.1,0.3,0.5,0.7,0.9\}$ on the evaluated tasks. For \method{}, we use the actual probe-output thresholds $\tau\in\{0.1,0.3,0.5,0.7,0.9\}$. Scores below the cutoff trigger the direct-answer prefill; otherwise they trigger the tool-use prefill. Thus, the columns denote quantile cutoffs for the baselines and probability thresholds for the probe, and do not necessarily represent matched call budgets. Increasing either cutoff favors skipping tools. Repeated verbalized-confidence operating points reflect tied scores.

\begin{table}[H]
\caption{Confidence-based controllers versus \method{} on Qwen3-4B-Instruct (soft prefill, 2{,}250 tasks, three runs). Each cell reports accuracy (\%) / mean calls per task. Column values are score quantiles $q$ for the three baselines and probe-output thresholds $\tau$ for \method{}. Compare the reported call counts when assessing accuracy at similar budgets.}
\label{tab:confidence_controllers}
\centering
\resizebox{\linewidth}{!}{%
\begin{tabular}{lccccc}
\toprule
Controller & $q$ or $\tau=0.1$ & 0.3 & 0.5 & 0.7 & 0.9 \\
\midrule
Self-ask & 88.6 / 0.949 & 87.1 / 0.757 & 85.5 / 0.620 & 81.4 / 0.497 & 77.5 / 0.361 \\
Verbalized confidence & 89.5 / 1.025 & 89.5 / 1.025 & 89.5 / 1.025 & 82.4 / 0.519 & 78.2 / 0.356 \\
First-token likelihood & 86.7 / 0.936 & 85.6 / 0.760 & 82.9 / 0.597 & 78.4 / 0.449 & 75.2 / 0.324 \\
\method{} & 91.1 / 0.985 & 90.4 / 0.759 & 88.5 / 0.582 & 84.8 / 0.456 & 74.7 / 0.292 \\
\bottomrule
\end{tabular}%
}
\end{table}

\paragraph{Results.}
At approximately 0.76 calls per task, \method{} achieves 90.4\% accuracy, compared with 87.1\% for self-ask and 85.6\% for first-token likelihood. At $\tau=0.5$, the probe achieves 88.5\% accuracy with 0.582 calls per task, exceeding the nearby self-ask and first-token operating points in accuracy while using fewer calls. At $\tau=0.7$, it also achieves higher accuracy with fewer calls than the verbalized-confidence controller at $q=0.7$. The most aggressive probe threshold yields the fewest calls but lower accuracy than the baselines' most aggressive listed points, illustrating the remaining accuracy--efficiency trade-off.

\newpage
\section{Necessity Supervision versus Difficulty Supervision}
\label{app:difficulty_supervision}

To examine whether the benchmark's difficulty tiers alone provide an adequate supervision signal, we train a second hidden-state probe using easy tasks as negative labels and hard tasks as positive labels. Medium tasks are excluded from this probe's training, leaving 600 easy/hard examples from the 900-example training split. The necessity probe uses all 900 examples labeled by the model's measured no-tool outcomes. Both probes use all-layer hidden-state features, standardization, and L2-regularized logistic regression; the difficulty probe uses inverse regularization strength $C=10^{-4}$. We evaluate both probes through the same soft-prefill pipeline on all 2{,}250 single-hop test tasks, including medium tasks, using Qwen3-4B-Instruct over three runs.

\begin{table}[H]
\caption{Necessity versus difficulty supervision on Qwen3-4B-Instruct. Each cell reports mean accuracy (\%) / mean calls per task over three runs. Both rows use actual probe-output thresholds $\tau$. Difficulty supervision excludes medium training examples; necessity supervision uses measured no-tool outcomes across all tiers.}
\label{tab:difficulty_supervision}
\centering
\resizebox{\linewidth}{!}{%
\begin{tabular}{lccccc}
\toprule
Probe & $\tau=0.1$ & $\tau=0.3$ & $\tau=0.5$ & $\tau=0.7$ & $\tau=0.9$ \\
\midrule
Necessity & 91.1 / 0.985 & 90.4 / 0.759 & 88.5 / 0.582 & 84.8 / 0.456 & 74.7 / 0.292 \\
Difficulty-only & 90.9 / 0.930 & 85.5 / 0.644 & 83.9 / 0.538 & 82.2 / 0.469 & 75.9 / 0.312 \\
\bottomrule
\end{tabular}%
}
\end{table}

\paragraph{Results and scope.}
At $\tau=0.7$, the necessity probe achieves 84.8\% accuracy with 0.456 calls per task, compared with 82.2\% accuracy and 0.469 calls for the difficulty-only probe. At other thresholds, the probes reach different operating points; neither row dominates every point in the other row. The results support the usefulness of model-specific necessity supervision beyond this easy-versus-hard control. They do not establish that the hidden-state probe is independent of surface difficulty cues, and the comparison also differs in training-set size and coverage of medium tasks.


\end{document}